\ifdefined\pdfminorversion\pdfminorversion=7\fi
\documentclass[onecolumn]{main}

\ifdefined\paperstyle
  \paperstyle{simple}
  \papercolor{navy}
\fi
\IfFileExists{lmodern.sty}{}{\microtypesetup{expansion=false}}

\usepackage{longtable}
\usepackage{listings}
\usepackage{wrapfig}
\usepackage{needspace}
\usepackage{float}
\tcbuselibrary{breakable,skins}

\ifdefined\paperstyle
  \newcommand{\colhead}[1]{\makebox[\linewidth][c]{#1}}
  \newcommand{\subfigref}[2]{Figure~\ref{#1}(\subref{#2})}

  \newcommand{\cellstack}[2]{\shortstack{#1\\[-1pt]#2}}
  \newcommand{\htmltabscale}{1}
  \newcommand{\ratioRx}{$R/x$}
  \newcommand{\ratioxR}{$x/R$}
  \newenvironment{contriblist}{\begin{itemize}[noitemsep,topsep=2pt,leftmargin=*]}{\end{itemize}}
\else
  \newcommand{\colhead}[1]{#1}
  \newcommand{\subfigref}[2]{Figure~\ref{#2}}
  
  \newcommand{\cellstack}[2]{#1 #2}
  \newcommand{\htmltabscale}{1.65}
  \newcommand{\ratioRx}{\textit{R}/\textit{x}}
  \newcommand{\ratioxR}{\textit{x}/\textit{R}}
  \newenvironment{contriblist}{\begin{itemize}[noitemsep,topsep=2pt]}{\end{itemize}}
\fi

\newlength{\fdetabw}

\definecolor{DELIVERHeader}{HTML}{EAF0F6}
\definecolor{DELIVERStripe}{HTML}{FAFBFD}
\definecolor{DELIVERTotal}{HTML}{F3F6FA}
\definecolor{DELIVERPromptBack}{HTML}{F8FAFC}
\definecolor{DELIVERPromptFrame}{HTML}{CCD7E2}
\definecolor{DELIVERPromptTitle}{HTML}{EAF0F6}

\title{FDE-Bench: Evaluating\\
End-to-End Delivery from Underspecified\\
Real-World Business Requests}

\ifdefined\paperstyle
\author[1,\ast]{Huaiming~Li}
\author[2,\ast]{Can~Huang}
\author[1]{Chufan~Wu}
\author[1]{Danyu~Liu}
\author[1]{Xiaomin~Yuan}
\author[1]{Qinze~Li}
\author[1]{Yui~Lo}
\author[1]{Wenhui~Bai}
\author[1]{Qianlong~Wang}
\author[1]{Zengle~Ge}
\author[1]{Xiyu~Yan}
\author[3]{Yuheng~Cheng}
\author[4]{Jinhui~Ren}
\author[4]{Guokai~Chen}
\author[5]{Jiaqun~Liu}
\author[6]{Haoran~Li}
\author[7]{Qianhui~Liu}
\author[1]{Annan~Li}
\author[1]{Jianmin~Wu}
\author[1]{Dawei~Yin}
\author[1,\dagger]{Dou~Shen}

\affiliation[1]{FM Agent Team, Baidu AI Cloud}
\affiliation[2]{Beijing University of Posts and Telecommunications}
\affiliation[3]{The Chinese University of Hong Kong}
\affiliation[4]{Tsinghua University}
\affiliation[5]{Peking University}
\affiliation[6]{University of the Chinese Academy of Sciences}
\affiliation[7]{Institute of Automation, Chinese Academy of Sciences}
\else
\author{%
  {Huaiming~Li\textsuperscript{1,\textasteriskcentered},
  Can~Huang\textsuperscript{2,\textasteriskcentered},
  Chufan~Wu\textsuperscript{1},
  Danyu~Liu\textsuperscript{1},
  Xiaomin~Yuan\textsuperscript{1},
  Qinze~Li\textsuperscript{1},
  Yui~Lo\textsuperscript{1},
  Wenhui~Bai\textsuperscript{1},
  Qianlong~Wang\textsuperscript{1},
  Zengle~Ge\textsuperscript{1},
  Xiyu~Yan\textsuperscript{1},
  Yuheng~Cheng\textsuperscript{3},
  Jinhui~Ren\textsuperscript{4},
  Guokai~Chen\textsuperscript{4},
  Jiaqun~Liu\textsuperscript{5},
  Haoran~Li\textsuperscript{6},
  Qianhui~Liu\textsuperscript{7},
  Annan~Li\textsuperscript{1},
  Jianmin~Wu\textsuperscript{1},
  Dawei~Yin\textsuperscript{1},
  Dou~Shen\textsuperscript{1,\textdagger}}\\[4pt]
  {\small
  \textsuperscript{1}FM Agent Team, Baidu AI Cloud\\
  \textsuperscript{2}Beijing University of Posts and Telecommunications\\
  \textsuperscript{3}The Chinese University of Hong Kong\\
  \textsuperscript{4}Tsinghua University\\
  \textsuperscript{5}Peking University\\
  \textsuperscript{6}University of the Chinese Academy of Sciences\\
  \textsuperscript{7}Institute of Automation, Chinese Academy of Sciences\\
  \textsuperscript{\textasteriskcentered}Equal contribution.\quad
  \textsuperscript{\textdagger}Corresponding author.}}
\fi

\abstract{
Real-world business requests often leave key requirements unstated because
frontline practitioners struggle to articulate rules they know from experience.
Forward-deployed engineers (FDEs) must elicit these requirements
through dialogue and iteratively optimize solutions until they meet customer
acceptance criteria.
Existing optimization benchmarks typically start from explicit objectives,
while clarification benchmarks often introduce information gaps into existing
tasks. We introduce FDE-Bench, the first benchmark built from real
customer projects to evaluate agents on end-to-end delivery tasks that require
interactive requirement clarification and iterative optimization.
Its 49 cases span combinatorial optimization and machine learning.
Clarification targets are distilled from discussions between frontline
practitioners and FDEs. Agents elicit missing requirements from a simulated
customer, and case-specific evaluators assess final artifacts against
acceptance criteria using customer-accepted FDE solutions as quality references.
Across 18 agent configurations, the highest mean delivery score is only
\textbf{0.5510}, compared with an FDE reference of \textbf{1.0}. The highest
delivery success rate (DSR), defined as the fraction of planned runs reaching this
reference, is only \textbf{6.12\%}. These results show that reliable FDE-level
end-to-end delivery remains out of reach for the evaluated agents on
underspecified real-world business tasks.
The benchmark project page is available at
\href{https://baidubce.github.io/FM-Agent/FDE-Bench/}{baidubce.github.io/FM-Agent/FDE-Bench}.
}

\begin{document}
\maketitle
\ifdefined\paperstyle
\begingroup
\renewcommand{\thefootnote}{\fnsymbol{footnote}}
\footnotetext[1]{Equal contribution.}
\footnotetext[2]{Corresponding author.}
\endgroup
\fi

\section{Introduction}
\label{sec:introduction}

\begin{figure*}[t]
    \centering
    \begin{subfigure}[t]{0.65\textwidth}
        \vspace{0pt}
        \centering
        \includegraphics[width=\linewidth]{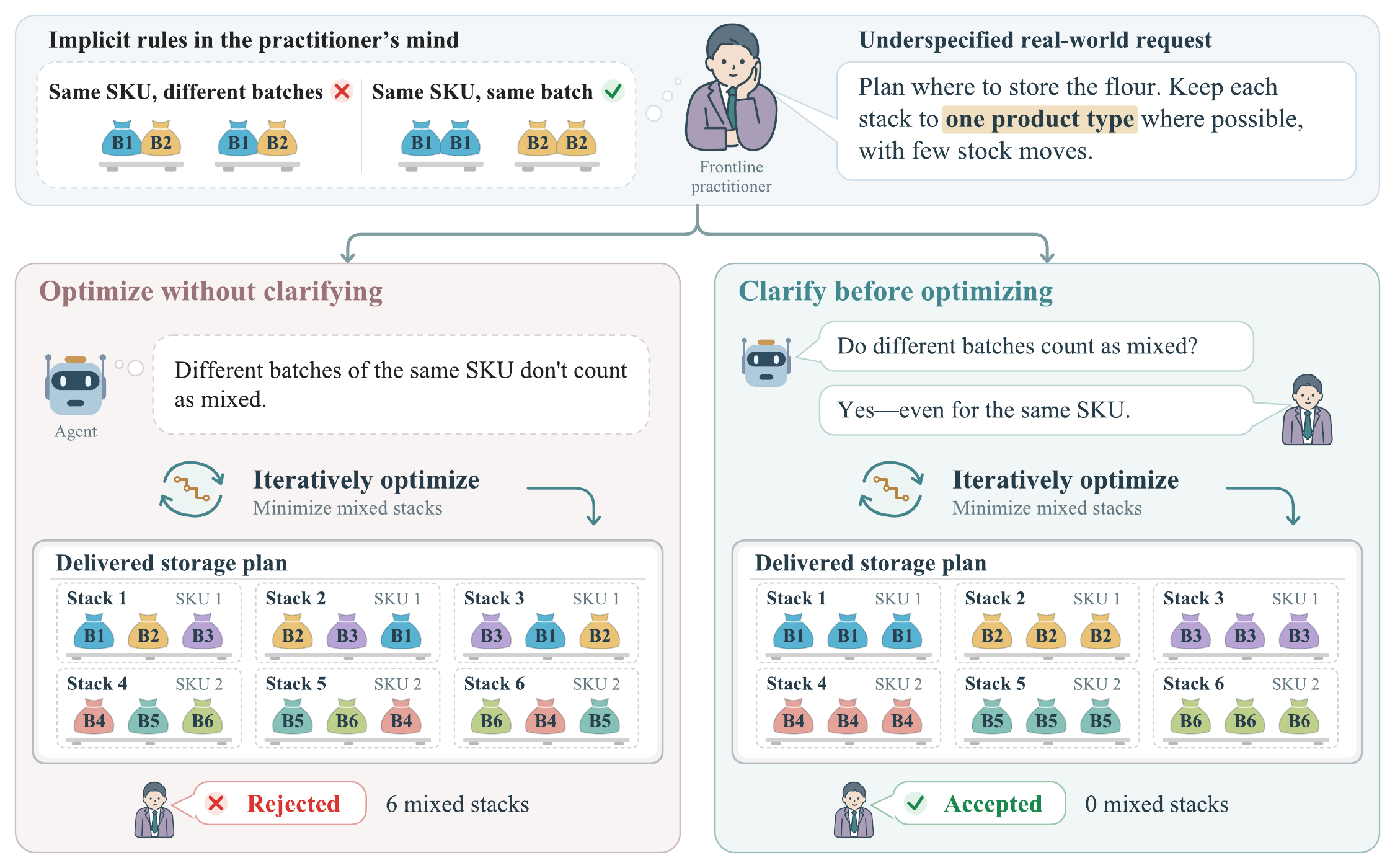}
        \caption{\textbf{A warehouse case}: clarify the mixing rule, then optimize.}
        \label{fig:motivation-case}
    \end{subfigure}\hfill
    \begin{subfigure}[t]{0.33\textwidth}
        \vspace{0pt}
        \centering
        \ifdefined\paperstyle
            \includegraphics[width=\linewidth]{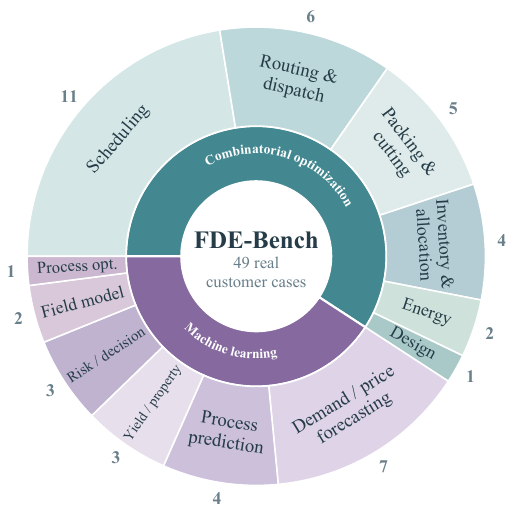}
        \else
            \includegraphics[width=\linewidth]{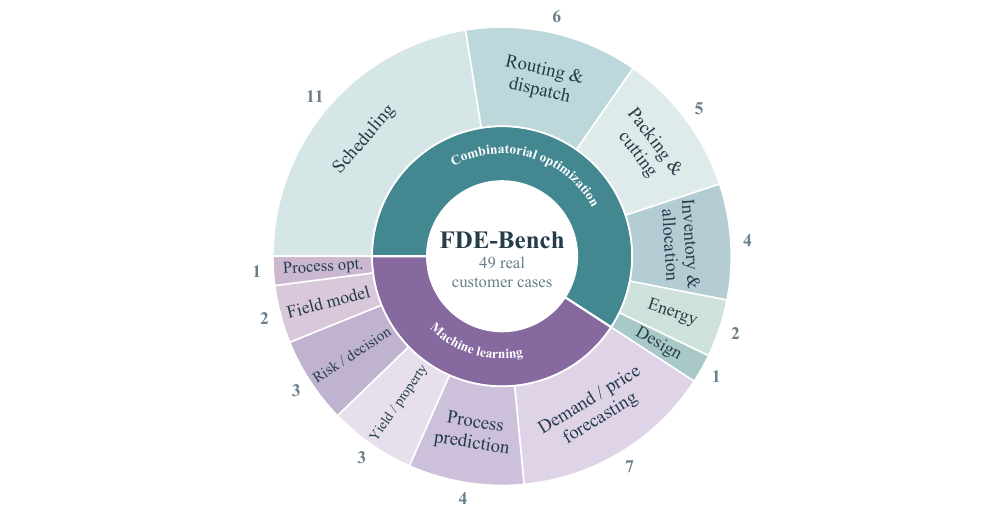}
        \fi
        \caption{\textbf{Overview of FDE-Bench:} 49 real customer cases
        across task families and subtypes.}
        \label{fig:bench-overview}
    \end{subfigure}
    \caption{\textbf{FDE-Bench: motivation and coverage.}
    (a) Clarifying mixing rules guides iterative warehouse optimization.
    (b) 49 real customer cases span two task families.}
    \label{fig:deliver-bench-motivation}
\end{figure*}

In real customer projects, frontline practitioners often rely on business
rules learned through experience but struggle to make them explicit in their
requests~\citep{nuseibeh2000requirements}.
Forward-deployed engineers (FDEs) must elicit these tacit requirements through
dialogue and iteratively optimize solutions until they meet the customer's
acceptance criteria~\citep{zowghi2005requirements}.
\subfigref{fig:deliver-bench-motivation}{fig:motivation-case}
illustrates a warehouse request to keep each stack to ``\textbf{one product type}''
where possible while avoiding unnecessary stock moves.  The request leaves
implicit that different batches of the same product (SKU) also count as mixed.
Grouping only by SKU can therefore leave mixed stacks under the customer's
criterion.  After clarifying this rule, the agent must still iteratively refine
storage assignments and moves to meet the delivery requirements.

Representative benchmarks do not yet fully capture this challenge.
Optimization benchmarks such as HeuriGym~\citep{chen2026heurigym} and
Frontier-Eng~\citep{chi2026frontier} provide explicit objectives and
constraints, leaving requirement acquisition outside their scope.
ClarEval~\citep{li2026clareval} perturbs programming instructions to create
omissions and ambiguities, while
HIL-Bench~\citep{trinh2026hilbench} introduces human-validated blockers
requiring interaction in coding and SQL tasks.  Their information gaps are
constructed within predefined tasks; we instead ask whether agents can elicit
requirements grounded in frontline practitioners' experience and iteratively
optimize solutions to meet real customers' delivery standards.

We introduce FDE-Bench, a benchmark of 49 cases from real customer
projects for evaluating agents from requirement clarification through
iterative optimization to final delivery.  As summarized in
\subfigref{fig:deliver-bench-motivation}{fig:bench-overview},
the cases cover 29
combinatorial-optimization and 20 machine-learning
tasks across diverse business settings.  Each case links the original request
and business data, requirements distilled from frontline practitioner--FDE
discussions, and a customer-accepted FDE solution.
Agents clarify requirements with a simulated customer, then iteratively
execute and improve their solutions.  After submission, case-specific
evaluators assess final artifacts against the project's acceptance criteria,
using the accepted FDE solution as the quality reference.
Separate clarification scores measure how well agents' questions cover the
annotated requirements.  Since performance depends on
both the model and its scaffold~\citep{kapoor2026holistic}, we build an extensible
evaluation harness to compare different configurations under a shared protocol.

We evaluate 18 agent configurations spanning seven model families and six
agent scaffolds.  Across configurations, the highest mean delivery score is
only 0.5510 against a customer-accepted FDE reference of 1.0.  The highest
delivery success rate (DSR)---the fraction of planned runs reaching that
reference---is only 6.12\%.  Broader requirement coverage in agents' questions
is associated with higher delivery quality.  Paired information comparisons
show higher scores with required clarification than with no customer answers,
but even complete business information leaves all compared configurations
below the FDE reference on average.  These results expose gaps in both
requirement acquisition and solution optimization.

Our contributions are:
\begin{contriblist}
    \item \textbf{A benchmark for end-to-end FDE delivery.}
    FDE-Bench is the first benchmark to couple interactive requirement
    clarification with iterative optimization on real customer projects,
    using customer-accepted FDE solutions as delivery references.
    \item \textbf{An extensible evaluation harness.}
    Backend adapters support new models and agent scaffolds; a shared protocol
    records interactions and evaluates submitted artifacts, with separate
    measures of requirement coverage and delivery quality.
    \item \textbf{A systematic study of current agents.}
    Across 18 configurations, we analyze representative delivery failures and
    the association between requirement coverage and delivery quality. Information ablations
    show that complete business information alone does not close the delivery gap.
\end{contriblist}

\section{Related Work}
\label{sec:related-work}

\newcommand{\delivercheck}{\ding{51}}
\newcommand{\delivercross}{\ding{55}}

\textbf{Optimization modeling and search.}
NL4Opt and OptiMUS study natural-language optimization modeling, while
OptiBench and OptMATH provide data for evaluating or training modeling
systems \citep{ramamonjison2023nl4opt,ahmaditeshnizi2024optimus,
yang2025optibench,lu2025optmathscalablebidirectionaldata}.  MIPLIB-NL extends
evaluation to industrial-scale mixed-integer problems \citep{li2026miplibnl}.
CO-Bench, HeuriGym, and ALE-Bench evaluate iterative algorithm search and
optimization \citep{sun2025cobench,chen2026heurigym,imajuku2025alebench}, while
ORAgentBench and Opti-Agent-Bench extend evaluation to workflows involving
problem formulation, file inspection, and execution
\citep{li2026oragentbench,fu2026optiagentbench}.  These works primarily assess
how agents formulate and solve given optimization tasks.  FDE-Bench
additionally evaluates whether agents can elicit unstated business requirements
through customer interaction and use them to guide optimization.

\textbf{Clarification and interactive agents.}
MINT evaluates multi-turn tool use with language feedback, while $\tau$-bench
studies customer interaction and policy compliance in API-based service tasks
\citep{wang2024mint,yao2024taubench}.  CLAMBER focuses on identifying ambiguous
information needs and generating clarifying questions \citep{zhang2024clamber}.
HIL-Bench introduces blockers into coding and SQL tasks, while ClarEval injects
ambiguities into programming instructions \citep{trinh2026hilbench,li2026clareval}.
Ambig-SWE evaluates clarification in software tasks~\citep{vijayvargiya2026ambigswe},
while \citet{pan2025whenbenchmarks} study code generation with interactive feedback.
SWE-RPG diagnoses implicit-requirement recovery from coding trajectories but
does not evaluate interactive user feedback \citep{zhou2026swerpg}.
FDE-Bench grounds clarification targets in discussions between
frontline practitioners and FDEs from real customer projects, then evaluates
delivery quality after agents clarify requirements and iteratively optimize
their solutions.

\textbf{Executable evaluation and expert references.}
SWE-bench and Terminal-Bench use executable verification to assess software
and command-line tasks \citep{jimenez2024swebench,merrill2026terminalbench}.
Frontier-Eng and EngDesign evaluate engineering solutions through task-specific
scoring or simulation \citep{chi2026frontier,guo2025engdesign}.  MLE-bench
measures machine-learning engineering performance, while RE-Bench compares
agents with human experts on ML research engineering tasks
\citep{chan2025mlebench,wijk2025rebench}.  FDE-Bench also uses executable
evaluation and reference solutions.  Its distinguishing feature is that the
initial request, clarification targets, and customer-accepted FDE reference
originate from the same customer project.  Agents must elicit initially
unstated requirements and optimize their deliverables against that project's
acceptance criteria.  Appendix~\ref{app:benchmark-comparison} provides a
comparison with representative benchmarks.

\section{FDE-Bench: Tasks and Construction}
\label{sec:benchmark}

\subsection{FDE Delivery Workflow}
\label{sec:case-collection}

\begin{figure}[t]
    \centering
    \includegraphics[width=0.85\linewidth]{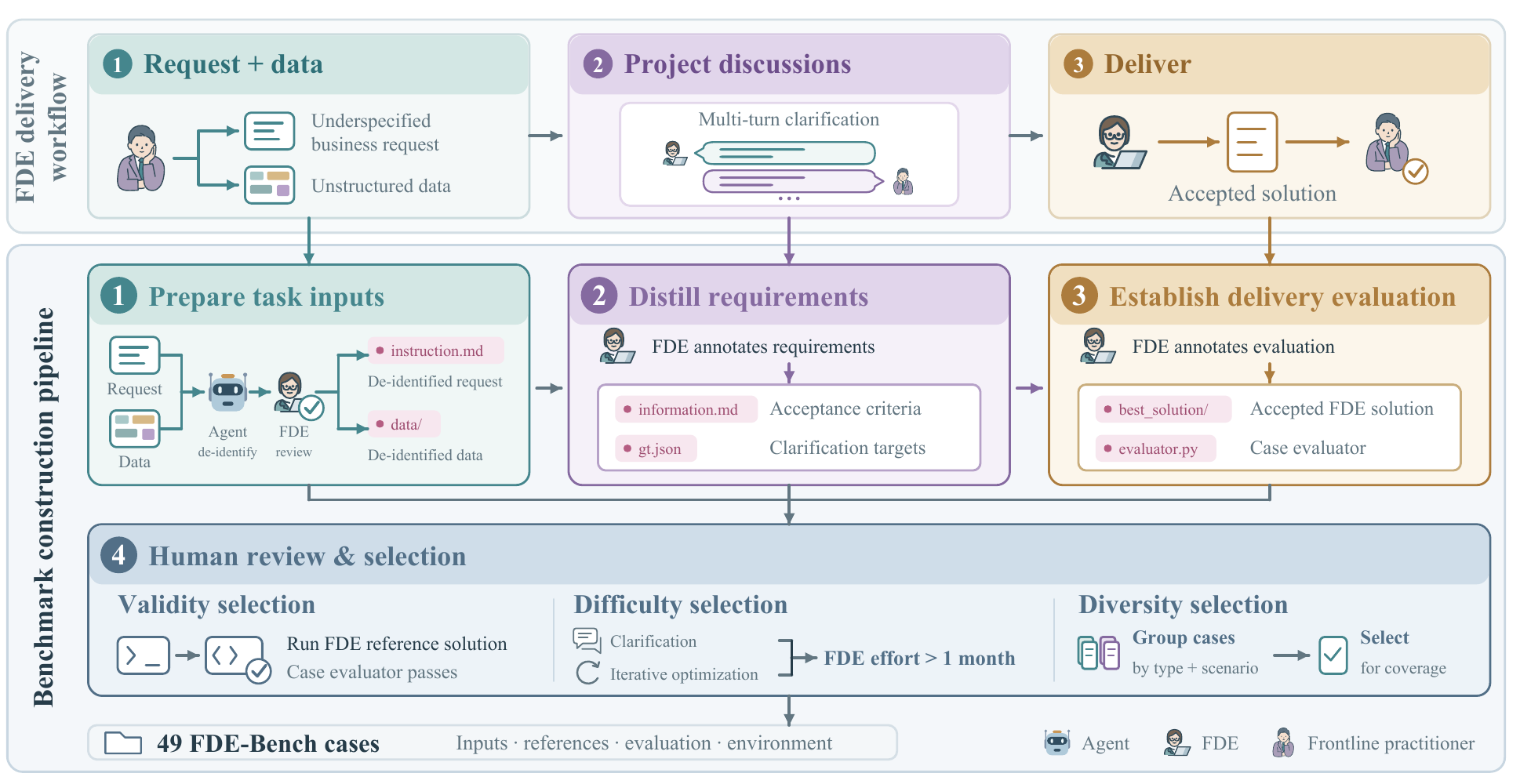}
    \caption{\textbf{FDE-Bench construction pipeline.}
    Real-project materials become agent-visible inputs, annotated requirements,
    case-specific evaluation, and a reviewed case package.}
    \label{fig:construction}
\end{figure}

Each case draws on the FDE delivery workflow of a real customer project
(Figure~\ref{fig:construction}). Frontline practitioners provide an
underspecified business request and business data. Through discussion, the
FDE elicits business rules and acceptance criteria, then iteratively optimizes
a solution until customer acceptance. The initial request and data, project
discussions, and accepted FDE solution from the same project supply task
inputs, requirement annotations, and delivery references, respectively.
The original discussion transcripts remain hidden from evaluated agents.

\subsection{Benchmark Construction Pipeline}
\label{sec:clarification-annotations}

\paragraph{Prepare task inputs.}
The initial request and business data are de-identified with agent assistance
and reviewed by an FDE. The process preserves the request's original meaning
and level of detail, applies replacements consistently across related files,
and retains numerical relationships and constraints needed for execution and
evaluation. The visible inputs are \texttt{instruction.md} and \texttt{data/};
the de-identification agent does not decide requirements or acceptance criteria.

\paragraph{Distill requirements.}
The FDE responsible for the original delivery annotates requirements
established in the customer--FDE discussions. We store complete business
context in \texttt{information.md} and clarification targets in
\texttt{gt.json}. A target is retained only if it cannot be resolved from
the initial request and visible data, and its resolution can affect
feasibility, the objective or target value, or customer acceptance.
Examples include the meaning of a data field, the scope of a constraint,
and the convention used to measure an objective. Each annotation records
the requirement and the information to clarify, without prescribing question
wording or dialogue order. The 49 cases contain 266 targets, or 5.43 per
case on average.

\paragraph{Establish delivery evaluation.}
Complete requirements and project acceptance criteria are translated into a
case-specific evaluator that checks the artifact schema, hard constraints, and
case-defined quality. The customer-accepted FDE deliverable supplies the
reference, normalized to 1.0. This is an achieved project standard rather than
a claim of global optimality; different algorithms are therefore judged against
a common set of requirements.

\paragraph{Human review and selection.}
Reviewers run each FDE reference solution with its evaluator and check
that the evaluation matches the documented business requirements and
acceptance criteria. They also review the FDE annotations using common
criteria across cases, checking targets against the evaluator and reference
artifact, removing duplicates, and confirming that each target has a
documented customer answer. Cases with incomplete materials or inconsistent
checks are excluded. To select projects involving sustained delivery work,
we retain those that required more than one month of FDE effort on
clarification and iterative optimization before customer acceptance.
Reviewers select cases across computational task types and business
settings to ensure coverage.

\subsection{Task Coverage and Specification}
\label{sec:case-corpus}
\label{sec:task-definition}

The release contains 29 combinatorial-optimization and 20 machine-learning
cases, spanning manufacturing, logistics, energy, retail, finance, and
engineering settings. Primary computational subtypes are listed in the case
inventory; scenario tags can overlap and are not mutually exclusive counts.

We represent each case as $\mathcal C=(u,D,I,G,E)$, comprising the initial
request $u$, business data $D$, complete business-information file $I$,
clarification targets $G$, and case-specific evaluator $E$. Each released
case package also includes an FDE reference deliverable and its reference
quality, an execution environment, and metadata. Under the main condition,
\texttt{Interact-Req}, only $u$ and $D$ are initially exposed in the solving
workspace; $I$, $G$, $E$, and the reference deliverable remain outside it.
Section~\ref{sec:protocol} specifies how agents interact with the simulated
customer and how their submitted artifacts and clarification questions are
evaluated.

\section{Evaluation Protocol}
\label{sec:protocol}

\begin{figure}[t]
    \centering
    \includegraphics[width=0.82\linewidth]{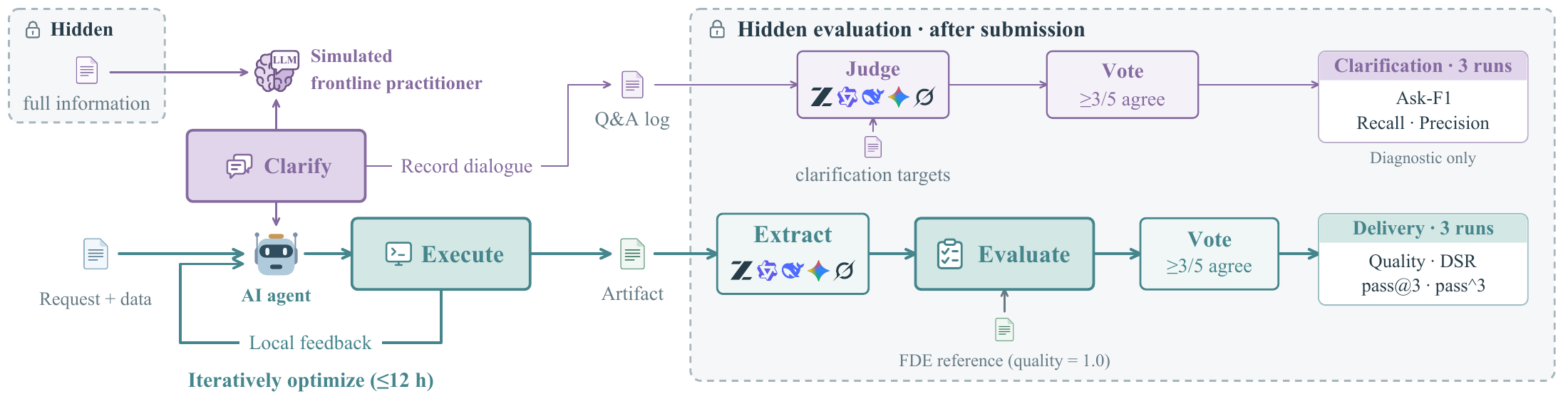}
    \caption{\textbf{FDE-Bench evaluation framework.} Agents clarify
    requirements and iteratively optimize deliverables. After submission,
    hidden evaluation extracts and scores artifacts against FDE references
    without online feedback; clarification coverage is assessed separately.}
    \label{fig:methods-overview}
\end{figure}

FDE-Bench evaluates two outputs of each run: the submitted solution file
(the \emph{artifact}) and the interaction log.  Case-specific evaluators
score delivery, while a separate judge panel assesses clarification
coverage (Figure~\ref{fig:methods-overview}).  Section~\ref{sec:condition-effects}
compares the main condition, \texttt{Interact-Req}, with no-answer and
full-information conditions.

\subsection{Evaluation Procedure}
\label{sec:interaction}

\paragraph{Customer interaction.}
Agents query a simulated customer for business information.
The customer uses the request $u$, complete business information $I$, and
dialogue history to answer, without access to clarification targets $G$.
We use GLM-5.3 with a fixed prompt and record exchanges.
Benchmark prompts and the native-tool, callback, and text-exchange
interfaces are provided with the benchmark.

\paragraph{Solving and submission.}
The \texttt{Interact-Req} prompt instructs agents to elicit missing
information before implementation, continuing until they consider the
information sufficient. Within the execution budget and interaction limits,
agents choose their questions, implement a candidate solution, execute it
locally, inspect the resulting files and diagnostics, and repeatedly revise
and re-execute the solution. They submit the final artifact only after this
local optimization loop. The main campaign allows 43,200 seconds (12 hours)
per solving run, including interaction, and at most 30 clarification rounds
with no separate question-count cap.
The hidden evaluator and artifact extractor run only after submission and
provide no score or evaluator feedback during this loop.

\paragraph{Extraction and case evaluation.}
\label{sec:execution-scoring}
After submission, extractors convert artifacts into the case evaluator's
input format while preserving the agent's solution and decisions. The
evaluator checks the case's hard constraints. Violations receive a delivery
score of zero; otherwise, it computes case-defined quality and normalizes
it to the customer-accepted FDE reference, which scores 1.0.
Up to five models---GLM-5.2, Qwen3.8-27B, DeepSeek-V4-Flash,
Gemini-3.6-Flash, and Grok-4.5---extract the artifact for scoring by the
same case evaluator.  Remaining routes are skipped once three scores agree
after rounding to four decimal places.  A score is recorded when at
least three routes agree.  The validity outcome is taken from an evaluation
supporting that score; this model-panel design is related to
prior work on LLM-assisted evaluation~\citep{zheng2023judging,liu2023g}.

\paragraph{Requirement coverage.}
We split recorded questions into units and use the same five models to assess
coverage of $G$. For each requirement, judges independently assess whether
answering at least one question would reveal it. Judges receive the requirement
and numbered questions, but not customer answers or requirement-category
metadata. Coverage is accepted when at least three of five judgments agree
($\geq 3/5$); otherwise the requirement is uncovered.

\subsection{Evaluation Metrics}
\label{sec:metrics}

FDE tasks require agents to elicit implicit business requirements and turn
them into a deliverable that meets the customer's standard.  We measure
clarification and delivery separately to examine whether effective
requirement acquisition translates into successful delivery.

\paragraph{Delivery metrics.}
In the main evaluation, each configuration runs three times on each of
49 cases. Quality first averages the three delivery scores within each case,
then averages across cases with equal weight. Delivery Success Rate (DSR)
is the fraction of the 147 runs reaching the FDE reference.
Let $s_{ir}$ denote the delivery score for repeat $r$ of case $i$. We define
\[
\mathrm{Quality}=\frac{1}{49}\sum_{i=1}^{49}
    \left(\frac{1}{3}\sum_{r=1}^{3}s_{ir}\right),
\qquad
\mathrm{DSR}=\frac{1}{147}\sum_{i=1}^{49}\sum_{r=1}^{3}
    \mathbf{1}[s_{ir}\geq1].
\]
Following the pass@k framing in code-generation
evaluation~\citep{chen2021evaluating}, we also report pass@3 and
pass\textasciicircum{}3: the fractions of cases with at least one and all
three valid runs, respectively. These measure validity across repeated runs;
DSR measures attainment of the FDE reference.

\paragraph{Clarification metrics.}
\label{sec:clarification-metrics}
Missing business requirements can lead an agent to solve the wrong
problem.  Recall ($R$) measures how fully its questions cover the reference
requirements, while Precision ($P$) measures how focused the questions are
on those requirements.  Their harmonic mean, Ask-F1, rewards broad coverage
with focused questioning, following the standard precision--recall terminology
in information retrieval~\citep{schutze2008introduction}.
For each run, let $G_{\mathrm{cov}}\subseteq G$ denote covered requirements,
$Q$ the set of question-unit indices, and $Q_{\mathrm{rel}}\subseteq Q$ the
indices matched to at least one covered requirement.  These three primary
metrics are
\[
R=\frac{|G_{\mathrm{cov}}|}{|G|},\qquad
P=\frac{|Q_{\mathrm{rel}}|}{|Q|},\qquad
\mathrm{Ask\text{-}F1}=\frac{2PR}{P+R}.
\]
Recall is question coverage: it measures whether a question could elicit a
reference requirement, not whether the customer stated it or the agent used it;
questions matching multiple requirements count once.

Efficient clarification should elicit business requirements with few customer
exchanges and limited question burden. First-turn coverage
($\mathrm{KQC}_1$) measures how much agents cover in the initial exchange,
while average turns to clarify (ATC) measures how early they address the
requirements eventually covered.  Unmatched-question count (Redund.)
captures the absolute number of questions unmatched to covered requirements.
Interpreted alongside Recall, these diagnostics characterize the trade-off
between requirement coverage and interaction effort.
Let $G_{\mathrm{cov},1}\subseteq G$ denote requirements covered in the first
interaction turn, and $G_{\mathrm{cov}}^{\star}\subseteq G_{\mathrm{cov}}$
those with a recorded matching question and turn.  For each requirement
$g\in G_{\mathrm{cov}}^{\star}$, let $t_g$ be its first covering turn,
counting from 1.  We define
\[
\begin{aligned}
\mathrm{KQC}_1=\frac{|G_{\mathrm{cov},1}|}{|G|},\qquad
\mathrm{ATC}=\frac{1}{|G_{\mathrm{cov}}^{\star}|}
    \sum_{g\in G_{\mathrm{cov}}^{\star}}t_g,\qquad
\mathrm{Redund.}=|Q|-|Q_{\mathrm{rel}}|.
\end{aligned}
\]
Clarification metrics are computed per run and averaged over runs for which
each metric is defined.

\subsection{Evaluation Reliability}

Two human experts independently audited stratified samples of retained
traces, covering 1,303 simulated-customer answers, 120 artifact extractions,
and 120 clarification items across all 18 configurations. Sampling
procedures, confidence intervals, and full audit results are provided in
Appendix~\ref{app:audits}.
For answer grounding, experts A and B marked 95.93\% and
90.48\% of answers as supported, respectively; their support labels agreed on
92.10\% of records (Cohen's $\kappa=0.385$~\citep{cohen1960coefficient}). For extraction fidelity,
on a sample enriched for runs with extraction issues, they judged 93.33\% and 91.00\% of assessable extractions faithful to the
submitted artifacts, respectively, with 91.67\% three-category agreement
(Cohen's $\kappa=0.7656$).
For clarification coverage, the model panel agreed with expert A on 94.74\%
($\kappa=0.8947$) and with B on 93.33\% ($\kappa=0.8667$), excluding
uncertain labels. These results support the reliability of the evaluation
pipeline on the audited subsets; the extraction $\kappa$ specifically measures
expert-label agreement, rather than extractor accuracy against an adjudicated
gold standard.  These checks reduce but do not eliminate extraction-related
score risk.

\section{Experiments}
\label{sec:experiments}

We evaluate 18 agent systems and ask three questions.  \textbf{RQ1}
(\S\ref{sec:main-result}): how close do agents come to the accepted FDE
deliverable?  \textbf{RQ2} (\S\ref{sec:failure-analysis}): which
clarification behaviors distinguish stronger systems?  \textbf{RQ3}
(\S\ref{sec:condition-effects}): how does delivery quality change with
access to business information?

\subsection{Experimental Setup}
\label{sec:exp-setup}

We evaluate scaffold--model systems: general scaffolds
\texttt{opencode} and \texttt{openhands}, each with seven models
(GPT-6-Astra, GLM-5.3, Qwen3.8-Max, Grok-4.6, DeepSeek-V4-Pro, Kimi-K3,
Gemini-3.1-Pro), and four native pairs (\texttt{codex},
\texttt{deepseek-harness}, \texttt{kimi}, \texttt{gemini}).  All main runs
use \texttt{Interact-Req} with a fixed simulated customer, a 12-hour budget,
and at most 30 clarification rounds; each system solves each of 49 cases
three times.  Paired contrasts use two-sided Wilcoxon
signed-rank tests~\citep{wilcoxon1945individual} on case means with Holm
correction~\citep{holm1979simple}. Settings and bootstrap intervals are
reported with the released results.

\subsection{RQ1: End-to-End Delivery}
\label{sec:main-result}

\begin{table}[htbp]
\centering
\fontsize{9}{11}\selectfont
\setlength{\tabcolsep}{2.5pt}
\setlength{\aboverulesep}{0.35ex}
\setlength{\belowrulesep}{0.35ex}
\renewcommand{\arraystretch}{1.05}
\caption{\textbf{End-to-end evaluation.} 49 cases, three runs per case,
\texttt{Interact-Req}. Quality is case-macro; other metrics are percentages.
Bold marks blockwise maxima before rounding.}
\label{tab:main-results}
\begin{tabularx}{\linewidth}{@{}l *{7}{>{\centering\arraybackslash}X}@{}}
\toprule
\rowcolor{DELIVERHeader}
\multicolumn{1}{@{}>{\columncolor{DELIVERHeader}[0pt][\tabcolsep]}l}{} &
\multicolumn{4}{c}{\textbf{Delivery $\uparrow$}} &
\multicolumn{3}{c}{\textbf{Clarification $\uparrow$}} \\
\noalign{\begingroup\color{DELIVERHeader}%
  \hrule height\dimexpr\aboverulesep+\cmidrulewidth+\belowrulesep\relax
  \vskip-\dimexpr\aboverulesep+\cmidrulewidth+\belowrulesep\relax
\endgroup}
\cmidrule(lr){2-5}\cmidrule(lr){6-8}
\rowcolor{DELIVERHeader}
\multicolumn{1}{@{}>{\columncolor{DELIVERHeader}[0pt][\tabcolsep]}l}{\textbf{Agent / model}} &
\colhead{\textbf{Quality}} &
\colhead{\textbf{DSR}} &
\colhead{\textbf{pass@3}} &
\colhead{\textbf{pass\textasciicircum{}3}} &
\colhead{\textbf{Recall}} &
\colhead{\textbf{Precision}} &
\multicolumn{1}{>{\columncolor{DELIVERHeader}[\tabcolsep][0pt]}c@{}}{\textbf{Ask-F1}} \\
\midrule
\rowcolor{DELIVERTotal}
\multicolumn{8}{@{}l}{\emph{General scaffold: \texttt{opencode}}} \\
GPT-6-Astra & \textbf{0.5459} & \textbf{5.44} & 91.84 & \textbf{81.63} & \textbf{83.64} & 36.53 & \textbf{48.82} \\
GLM-5.3 & 0.5413 & 1.36 & 91.84 & 77.55 & 66.83 & \textbf{39.70} & 47.46 \\
Qwen3.8-Max & 0.5387 & 2.72 & \textbf{95.92} & 73.47 & 70.56 & 35.41 & 45.11 \\
Grok-4.6 & 0.5221 & 2.72 & 87.76 & 79.59 & 68.19 & 38.30 & 46.42 \\
DeepSeek-V4-Pro & 0.4986 & 1.36 & 91.84 & 71.43 & 55.08 & 37.93 & 42.73 \\
Kimi-K3 & 0.4187 & 0.00 & 91.84 & 53.06 & 55.03 & 37.60 & 43.27 \\
Gemini-3.1-Pro & 0.3873 & 0.00 & 89.80 & 59.18 & 47.10 & 39.48 & 40.38 \\
\addlinespace[2pt]
\rowcolor{DELIVERTotal}
\multicolumn{8}{@{}l}{\emph{General scaffold: \texttt{openhands}}} \\
GPT-6-Astra & \textbf{0.5302} & 4.76 & \textbf{93.88} & 67.35 & \textbf{84.67} & 36.80 & \textbf{49.50} \\
GLM-5.3 & 0.5274 & 2.04 & 91.84 & 67.35 & 67.12 & 36.79 & 45.48 \\
Grok-4.6 & 0.5159 & 0.68 & 91.84 & \textbf{71.43} & 69.91 & 36.95 & 45.91 \\
DeepSeek-V4-Pro & 0.5104 & 0.68 & \textbf{93.88} & \textbf{71.43} & 65.24 & 36.41 & 44.92 \\
Qwen3.8-Max & 0.5071 & \textbf{6.12} & \textbf{93.88} & 65.31 & 75.37 & 30.89 & 42.24 \\
Gemini-3.1-Pro & 0.4571 & 0.00 & 87.76 & 65.31 & 45.45 & \textbf{37.77} & 39.25 \\
Kimi-K3 & 0.4119 & 0.68 & 85.71 & 57.14 & 57.94 & 35.39 & 41.34 \\
\addlinespace[2pt]
\rowcolor{DELIVERTotal}
\multicolumn{8}{@{}l}{\emph{Native scaffolds}} \\
\mbox{\texttt{codex} + GPT-6-Astra} & \textbf{0.5510} & \textbf{4.08} & \textbf{93.88} & \textbf{75.51} & \textbf{84.70} & 37.87 & \textbf{50.12} \\
\mbox{\texttt{deepseek-harness} + DeepSeek-V4-Pro} & 0.5271 & 1.36 & 91.84 & \textbf{75.51} & 61.79 & \textbf{39.32} & 45.93 \\
\mbox{\texttt{kimi} + Kimi-K3} & 0.4203 & 0.68 & 89.80 & 59.18 & 55.69 & 36.69 & 42.90 \\
\mbox{\texttt{gemini} + Gemini-3.1-Pro} & 0.4037 & 0.00 & 87.76 & 57.14 & 44.94 & 38.82 & 39.37 \\
\bottomrule
\end{tabularx}
\end{table}

\paragraph{FDE-level delivery remains rare.}
Across the 18 systems, the highest mean Quality is only 0.5510
(\texttt{codex} + GPT-6-Astra), and the highest DSR is 6.12\%
(Table~\ref{tab:main-results}). Only 8 of 49 cases have any reference-level
run, and in 23 the best system averages below 0.8.
Four systems, including all three Gemini-3.1-Pro systems, reach the
reference in none of their 147 runs, and no ML run reaches it in any system.

\paragraph{Validity varies across repeated runs.}
Across systems, 85.7--95.9\% of cases have at least one valid repeat among
three attempts (pass@3). However, pass\textasciicircum{}3 trails pass@3 for
every system by 8.2--38.8 percentage points (mean 23.0): for a typical
system, 23\% of cases are valid in some runs but not in all three.
Of the 548 zero-score runs, 78.5\% record a constraint or feasibility
violation and 1.5\% a schema error (Appendix~\ref{app:zero-score-analysis}):
agents submit well-formed artifacts that break business rules.

\paragraph{Runtime and case difficulty.}
Across systems, mean run time ranges from 0.3 to 5.5 hours, with no
significant association with Quality (Spearman $\rho=0.18$,
$p=0.47$)~\citep{spearman1961proof}. Case-level Quality rankings are similar
across systems (mean pairwise Spearman $\rho=0.71$).

\paragraph{The model matters more than the scaffold.}
With the scaffold fixed, Quality spans 0.159 across models on
\texttt{opencode} and 0.118 on \texttt{openhands}; with the model fixed, the
two general scaffolds differ by at most 0.032 for six models
(Gemini-3.1-Pro: 0.387 vs.\ 0.457).  Native scaffolds add 0.005 and 0.017
for GPT-6-Astra and DeepSeek-V4-Pro, while \texttt{gemini} trails
\texttt{openhands} by 0.053.

\subsection{RQ2: Clarification and Delivery}
\label{sec:failure-analysis}

\begin{wrapfigure}[13]{r}{0.50\linewidth}
\vspace{2pt}
\centering
\captionsetup{skip=4pt}
\includegraphics[width=\linewidth]{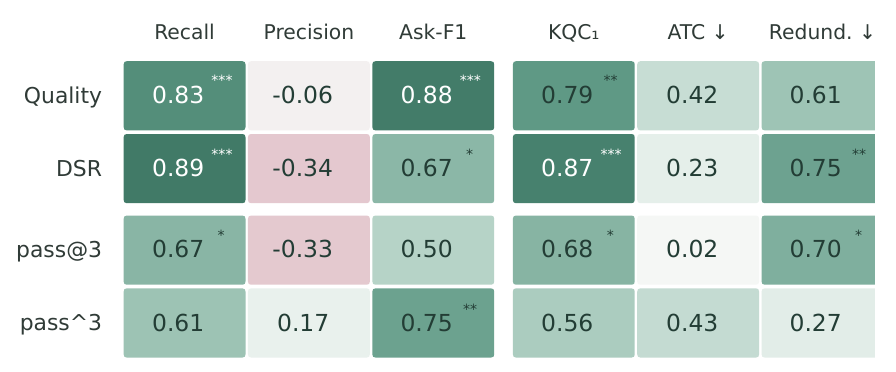}
\caption{\textbf{Clarification--delivery correlations.} Cells report rank
correlations across 18 configurations. Stars mark Holm-adjusted
two-sided $p$-values over 24 comparisons; down arrows denote lower-is-better
diagnostics.}
\label{fig:clarification-delivery-correlation}
\end{wrapfigure}

Figure~\ref{fig:clarification-delivery-correlation} correlates clarification
and delivery metrics across the 18 systems.

\Needspace{14\baselineskip}

\paragraph{Stronger systems cover more requirements; they do not ask more
precisely.}
Recall ranges from 44.9\% to 84.7\%, while Precision stays at 30.9--39.7\%;
every system spends 60--69\% of its question units on questions that match no
covered requirement.  Recall and
Ask-F1 track Quality ($\rho=0.83$ and $0.88$, Holm $p<0.001$), and Recall
tracks DSR ($\rho=0.89$), whereas Precision tracks no delivery metric
(Holm $p=1.0$).  Within systems and cases, higher-Recall repeats score higher
in 45\% and lower in 37\% of 626 cells (mean $+0.031$, Wilcoxon $p=0.012$).
Coverage of each of the four requirement domains also correlates with
\mbox{Quality} ($\rho=0.81$--$0.83$); domain-level results and question-type shares
are detailed in Appendix~\ref{app:clarification-types}.
These observational comparisons do not isolate the causal effect of
clarification.

\paragraph{Requirement coverage is concentrated in the first round.}
Average turns to clarify is 1.01--1.16, and first-turn coverage is 91--99\%
of final Recall. Under \texttt{Interact-Req}'s instruction to clarify before
implementation, most covered requirements are addressed in the initial
exchange.

\paragraph{More questions do not consistently improve delivery.}
Unmatched-question count correlates with DSR ($\rho=0.75$, Holm $p=0.006$):
systems that ask more cover more requirements and also ask more unmatched
questions. On the five most
requirement-dense cases, mandatory rather than optional clarification raises
question count for all eight tested systems and Recall for seven, but Quality
for only five; the changes
range from $-0.125$ to $+0.196$ and none is significant (Holm $p\ge0.50$).

\paragraph{Case study: asking about a rule does not ensure compliance.}
In case 10 (retail replenishment), \texttt{opencode} + GPT-6-Astra asks
whether same-city stock must be used before supplier purchases.
Its local validator reports \texttt{pass}, but the submitted plan purchases
an item from a supplier while the local warehouse still holds 327 units.
The hidden evaluator identifies violations of this priority rule at eight
stores, and the run receives Quality 0
(Appendix~\ref{app:case-study-019}). This run illustrates a gap between
asking about a business rule and enforcing it in the final deliverable.

\subsection{RQ3: Information Ablation}
\label{sec:condition-effects}
\label{sec:conditions}

Table~\ref{tab:condition-effects} compares \texttt{Hidden} (request and data
only), \texttt{Interact-Req} (main runs reused), and \texttt{Full} (complete
business information, no interaction) on 20 cases and eight systems; it
reports paired differences:
$\Delta_{\rm req}=\texttt{Interact-Req}-\texttt{Hidden}$ and
$\Delta_{\rm remain}=\texttt{Full}-\texttt{Interact-Req}$.

\begingroup
\setlength{\intextsep}{5pt}
\begin{table}[htbp]
\centering
\fontsize{9}{11}\selectfont
\setlength{\tabcolsep}{2.5pt}
\setlength{\aboverulesep}{0.35ex}
\setlength{\belowrulesep}{0.35ex}
\renewcommand{\arraystretch}{1.0}
\caption{\textbf{Information ablation.} Case-macro Quality on 20 paired
cases (three runs/condition). Bold marks maxima. Intervals and adjusted tests are in
Appendix~\ref{app:condition-statistics}.}
\label{tab:condition-effects}
\label{tab:conditions}
\begin{tabularx}{\linewidth}{@{}l *{5}{>{\centering\arraybackslash}X}@{}}
\toprule
\rowcolor{DELIVERHeader}
\multicolumn{1}{@{}>{\columncolor{DELIVERHeader}[0pt][\tabcolsep]}l}{\textbf{Agent / model}} &
\textbf{Hidden} & \colhead{\textbf{Interact-Req}} & \textbf{Full} &
$\boldsymbol{\Delta_{\rm req}}$ &
\multicolumn{1}{>{\columncolor{DELIVERHeader}[\tabcolsep][0pt]}c@{}}{$\boldsymbol{\Delta_{\rm remain}}$} \\
\midrule
\rowcolor{DELIVERTotal}
\multicolumn{6}{@{}l}{\emph{General scaffold: \texttt{opencode}}} \\
\texttt{opencode} + GPT-6-Astra & \textbf{0.1803} & \textbf{0.5800} & \textbf{0.7461} & \textbf{+0.3997} & +0.1662 \\
\texttt{opencode} + DeepSeek-V4-Pro & 0.1423 & 0.5063 & 0.6821 & +0.3640 & +0.1758 \\
\texttt{opencode} + Gemini-3.1-Pro & 0.1218 & 0.2912 & 0.4886 & +0.1694 & \textbf{+0.1974} \\
\texttt{opencode} + Kimi-K3 & 0.1244 & 0.3694 & 0.4488 & +0.2450 & +0.0794 \\
\addlinespace[2pt]
\rowcolor{DELIVERTotal}
\multicolumn{6}{@{}l}{\emph{Native scaffolds}} \\
\texttt{codex} + GPT-6-Astra & 0.1980 & \textbf{0.5980} & \textbf{0.7946} & \textbf{+0.4000} & +0.1966 \\
\texttt{deepseek-harness} + DeepSeek-V4-Pro & \textbf{0.1998} & 0.5642 & 0.7339 & +0.3644 & +0.1697 \\
\texttt{gemini} + Gemini-3.1-Pro & 0.1231 & 0.3763 & 0.5184 & +0.2532 & +0.1420 \\
\texttt{kimi} + Kimi-K3 & 0.1498 & 0.3693 & 0.7184 & +0.2195 & \textbf{+0.3490} \\
\bottomrule
\end{tabularx}
\end{table}
\endgroup

\paragraph{Interaction closes 39--76\% of the score gap.}
Under \texttt{Hidden}, systems score 0.12--0.20; \texttt{Interact-Req} adds
$+0.17$--$+0.40$ (significant for all eight, Holm over 24 tests), recovering
64--76\% of the \texttt{Hidden}-to-\texttt{Full} Quality gap for six systems, 46\% for
\texttt{opencode} + Gemini-3.1-Pro, and 39\% for \texttt{kimi} + Kimi-K3.
\texttt{Full} exceeds \texttt{Interact-Req} for all eight ($+0.08$--$+0.35$;
significant for five), although it removes interaction.

\paragraph{Full information leaves a substantial delivery gap.}
Under \texttt{Full}, mean Quality ranges from 0.449 to 0.795, leaving all
eight systems below the FDE reference of 1.0. Providing complete business
information improves delivery but is insufficient to reach the reference
on average.

\begingroup
\setlength{\intextsep}{2pt}
\begin{figure}[!htbp]
\centering
\includegraphics[width=\linewidth]{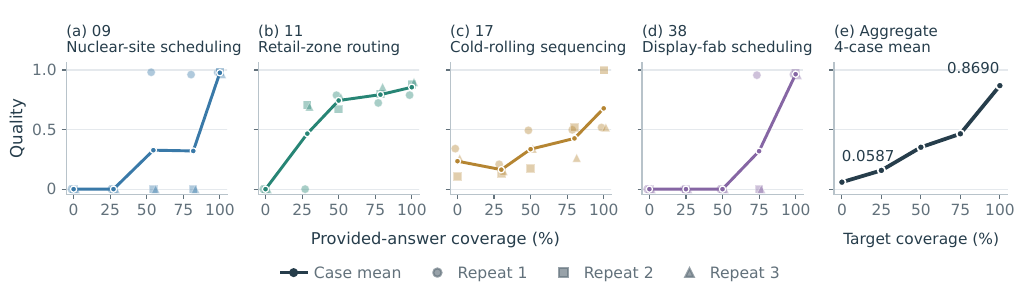}
\caption{\textbf{Oracle information coverage.} Codex + GPT-6-Astra receives
nested answer subsets; panels (a)--(d) show four cases and (e) their macro mean.}
\label{fig:oracle-coverage}
\end{figure}
\endgroup

\label{sec:oracle-coverage}
Oracle returns are nonlinear and case-dependent. We give \texttt{codex} +
GPT-6-Astra nested random subsets of the registered answers on four cases,
without interaction. At 100\% coverage, agents receive all registered answers,
not the complete business-information file used by \texttt{Full}.
Mean Quality rises from 0.059 at
0\% coverage to 0.352 at 50\% and 0.869 at 100\%; the last quarter of answers
yields half of the gain.  Case 38 scores zero in every run up to 50\%
coverage and 0.965 at full coverage, and at intermediate coverage cases 09
and 38 obtain a nonzero score in one of three repeats and score zero in the others
(Figure~\ref{fig:oracle-coverage}).  A few unresolved requirements can void
an otherwise reasonable solution, which is consistent with RQ1: the highest
Recall is 84.7\%, yet no system exceeds 6.12\% DSR.

\section{Conclusion}
\label{sec:conclusion}

We introduced FDE-Bench to evaluate agents' end-to-end delivery from
underspecified real-world business requests. Its 49 cases use customer-accepted
FDE solutions as delivery references. Across 18 configurations, the highest
mean Quality is only 0.5510 against an FDE reference of 1.0. The highest
delivery success rate is only 6.12\%. Broader requirement coverage is associated with
higher delivery quality, but even complete business information leaves all
compared configurations below the FDE reference on average. Reliable
FDE-level delivery therefore requires progress in both requirement
clarification and solution optimization.
FDE-Bench provides a shared testbed for comparing new agent systems and
examining whether better requirement elicitation translates into better
delivery.

\section{Limitations and Future Work}
\label{sec:limitations}

FDE-Bench evaluates delivery with stable business requirements and fixed
datasets. The simulated customer does not revise its requirements in response
to intermediate solutions, whereas real customers may reveal or change
expectations during development~\citep{zowghi2005requirements}. The benchmark
therefore does not yet assess agents' ability to maintain delivery quality
as production data and customer requirements evolve.
Coverage is limited to 49 real customer projects in combinatorial
optimization and machine learning. These cases do not cover all FDE settings.

We plan to expand FDE-Bench with additional real customer projects across
more domains, including scientific and engineering applications. Selection
will prioritize task types and business settings that are underrepresented
in the current release. New cases will undergo the same requirement review
and reference-solution checks. Future
evaluations should incorporate continuously updated production data,
evolving requirements, and human customers, testing both sustained delivery
quality and the realism of simulated interactions. Evaluation in these
settings should assess successive deliverables against the customer's current
acceptance criteria and track how agents respond to requirement changes.

\clearpage

\bibliographystyle{unsrtnat}
\bibliography{references}

\clearpage
\appendix
\section*{Appendix}
\raggedbottom
\section{Additional Evaluation Details}
\label{app:details}

\begin{tcolorbox}[colback=DELIVERStripe,colframe=DELIVERHeader,
  boxrule=0.4pt,arc=1pt,left=8pt,right=8pt,top=5pt,bottom=5pt,
  before skip=8pt,after skip=12pt]
\small
\renewcommand{\arraystretch}{1.2}
\begin{tabularx}{\linewidth}{@{}>{\bfseries\raggedright\arraybackslash}p{0.23\linewidth} >{\raggedright\arraybackslash}X@{}}
Positioning & Benchmark comparison (\ref{app:benchmark-comparison}). \\
Cases & Case map, scoring references, and clarification points
(\ref{app:case-inventory}). \\
Protocol & Delivery and process metrics and runtime settings
(\ref{app:delivery-metrics}). \\
Results & Full results (\ref{app:full-results}), clarification types
(\ref{app:clarification-types}), ablations (\ref{app:ablation-audit}), and
zero scores with a case study (\ref{app:zero-score-analysis}). \\
Reliability & Customer, judge, and extraction audits (\ref{app:audits}). \\
Prompts & Agent, customer, extractor, and judge templates (\ref{app:prompts}). \\
\end{tabularx}
\end{tcolorbox}

\subsection{Benchmark comparison}
\label{app:benchmark-comparison}

Table~\ref{tab:related-work-comparison} compares FDE-Bench with representative
benchmarks and systems (Section~\ref{sec:related-work}).  Requirements denotes
implicit business requirements annotated from customer--engineer discussions
in real delivery projects.  Reference denotes a customer-accepted project
solution used as a delivery-quality baseline.  These definitions concern the
provenance of the requirements and reference; RE-Bench, for example, provides
human expert comparisons and reference solutions, but not customer-accepted
project deliverables.

\begin{table}[h]
\centering
\footnotesize
\setlength{\tabcolsep}{2.5pt}
\renewcommand{\arraystretch}{1.15}
\caption{\textbf{Comparison with representative benchmarks and systems.}
\delivercheck{}: primary design feature; \delivercross{}: not a primary design
feature. Requirements and Reference are defined above. Clarification:
evaluation of clarifying questions or the use of user feedback; Execution:
task outcomes verified through code or tool execution.}
\label{tab:related-work-comparison}
\begin{tabularx}{\textwidth}{@{}>{\raggedright\arraybackslash}p{0.30\textwidth} >{\raggedright\arraybackslash}p{0.15\textwidth} *{4}{>{\centering\arraybackslash}X}@{}}
\toprule
\rowcolor{DELIVERHeader}
\textbf{Benchmarks / systems} & \textbf{Task type} & \textbf{Requirements} &
\textbf{Clarification} & \textbf{Execution} & \textbf{Reference} \\
\midrule
\textsc{MIPLIB-NL}~\citep{li2026miplibnl} & Optimization modeling & \delivercross{} & \delivercross{} & \delivercheck{} & \delivercross{} \\
\textsc{OptiMUS}~\citep{ahmaditeshnizi2024optimus} & Optimization modeling & \delivercross{} & \delivercross{} & \delivercheck{} & \delivercross{} \\
\textsc{OptiBench}~\citep{yang2025optibench} & Optimization modeling & \delivercross{} & \delivercross{} & \delivercheck{} & \delivercross{} \\
\textsc{CO-Bench}~\citep{sun2025cobench} & CO search & \delivercross{} & \delivercross{} & \delivercheck{} & \delivercross{} \\
\textsc{HeuriGym}~\citep{chen2026heurigym} & CO search & \delivercross{} & \delivercross{} & \delivercheck{} & \delivercross{} \\
\textsc{ALE-Bench}~\citep{imajuku2025alebench} & CO search & \delivercross{} & \delivercross{} & \delivercheck{} & \delivercross{} \\
\textsc{ORAgentBench}~\citep{li2026oragentbench} & End-to-end OR & \delivercross{} & \delivercross{} & \delivercheck{} & \delivercross{} \\
\textsc{Opti-Agent-Bench}~\citep{fu2026optiagentbench} & End-to-end OR & \delivercross{} & \delivercross{} & \delivercheck{} & \delivercross{} \\
\textsc{MINT}~\citep{wang2024mint} & Interactive agents & \delivercross{} & \delivercheck{} & \delivercheck{} & \delivercross{} \\
\textsc{$\tau$-bench}~\citep{yao2024taubench} & Interactive agents & \delivercross{} & \delivercheck{} & \delivercheck{} & \delivercross{} \\
\textsc{CLAMBER}~\citep{zhang2024clamber} & Clarification/HITL & \delivercross{} & \delivercheck{} & \delivercross{} & \delivercross{} \\
\textsc{HIL-Bench}~\citep{trinh2026hilbench} & Clarification/HITL & \delivercross{} & \delivercheck{} & \delivercheck{} & \delivercross{} \\
\textsc{ClarEval}~\citep{li2026clareval} & Clarification/HITL & \delivercross{} & \delivercheck{} & \delivercheck{} & \delivercross{} \\
\textsc{SWE-RPG}~\citep{zhou2026swerpg} & Requirement diagnosis & \delivercross{} & \delivercross{} & \delivercheck{} & \delivercross{} \\
\textsc{SWE-bench}~\citep{jimenez2024swebench} & Execution agents & \delivercross{} & \delivercross{} & \delivercheck{} & \delivercross{} \\
\textsc{Terminal-Bench}~\citep{merrill2026terminalbench} & Execution agents & \delivercross{} & \delivercross{} & \delivercheck{} & \delivercross{} \\
\textsc{Frontier-Eng}~\citep{chi2026frontier} & Execution agents & \delivercross{} & \delivercross{} & \delivercheck{} & \delivercross{} \\
\textsc{EngDesign}~\citep{guo2025engdesign} & Execution agents & \delivercross{} & \delivercross{} & \delivercheck{} & \delivercross{} \\
\textsc{MLE-bench}~\citep{chan2025mlebench} & ML/R\&D & \delivercross{} & \delivercross{} & \delivercheck{} & \delivercross{} \\
\textsc{RE-Bench}~\citep{wijk2025rebench} & ML/R\&D & \delivercross{} & \delivercross{} & \delivercheck{} & \delivercross{} \\
\rowcolor{blue!8}
\textbf{FDE-Bench (Ours)} & CO + ML & \delivercheck{} & \delivercheck{} & \delivercheck{} & \delivercheck{} \\
\bottomrule
\end{tabularx}
\end{table}
\subsection{Cases and clarification points}
\label{app:case-inventory}

\paragraph{Case map.}
\begin{figure}[H]
\centering
\includegraphics[width=\linewidth]{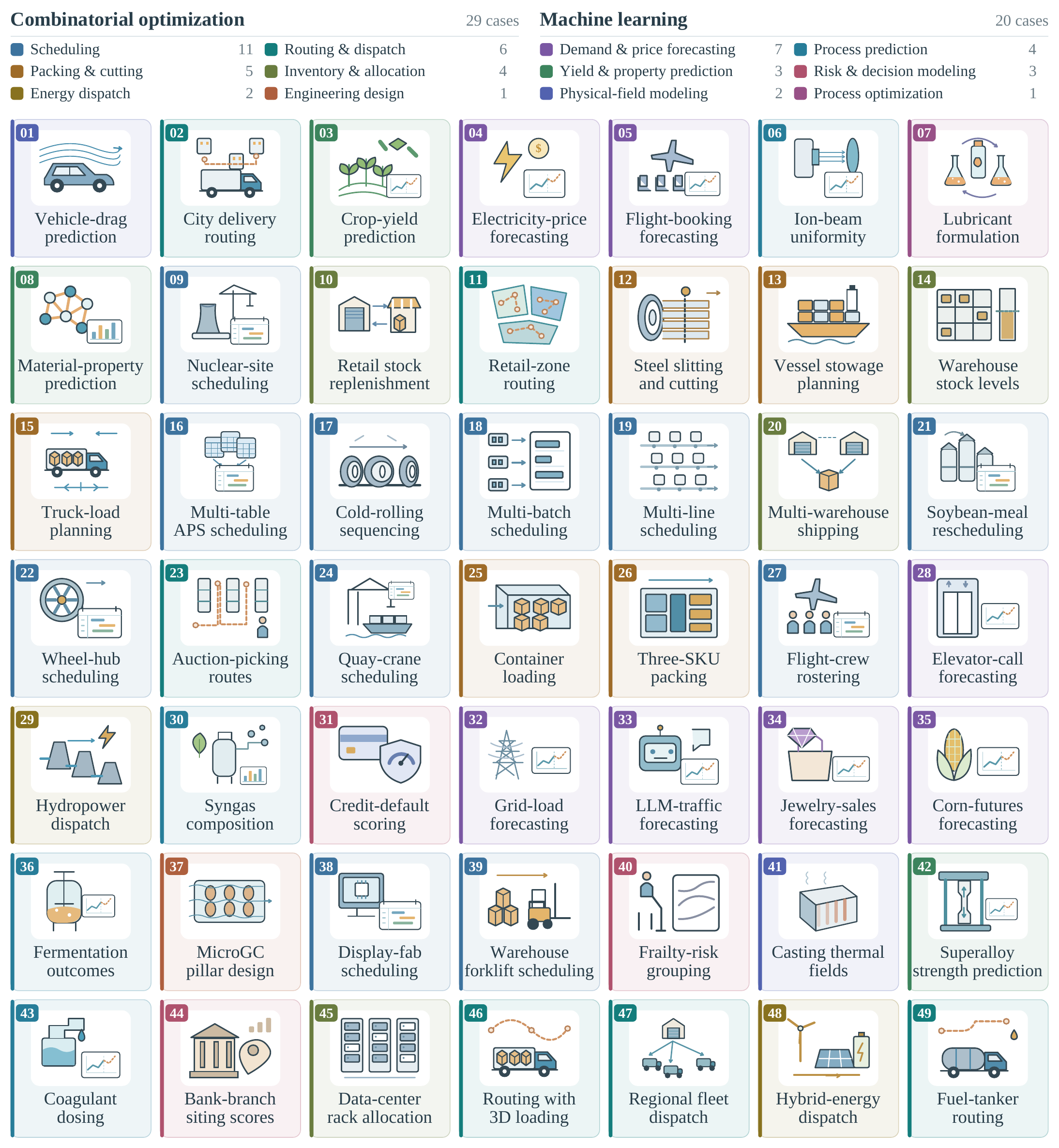}
\caption{\textbf{Case map.} The 49 cases by task family and subtype, numbered 01--49.}
\label{fig:case-map}
\end{figure}

Each case has one primary subtype for reporting; Figure~\ref{fig:case-map}
gives the counts.
\clearpage
\paragraph{Scoring references.}\label{app:case-scoring-reference}
Table~\ref{tab:case-evaluation-reference} lists each case's raw objective,
hard constraints, normalization direction, and FDE reference value $R$.

\begingroup
\scriptsize\setlength{\fdetabw}{\htmltabscale\linewidth}\setlength{\tabcolsep}{2pt}\setlength{\LTleft}{0pt plus 1fill}\setlength{\LTright}{0pt plus 1fill}\renewcommand{\arraystretch}{1.08}
\begin{longtable}{@{}>{\centering\arraybackslash}p{0.037\fdetabw} >{\raggedright\arraybackslash}p{0.178\fdetabw} >{\centering\arraybackslash}p{0.052\fdetabw} >{\raggedright\arraybackslash}p{0.246\fdetabw} >{\raggedright\arraybackslash}p{0.200\fdetabw} >{\centering\arraybackslash}p{0.121\fdetabw} >{\centering\arraybackslash}p{0.105\fdetabw}@{}}
\caption{\textbf{Case-level evaluator reference.} Numbering follows Figure~\ref{fig:case-map}.}\label{tab:case-evaluation-reference}\\
\toprule
\textbf{No.} & \textbf{Case} & \textbf{Fam.} & \textbf{Raw objective} & \textbf{Hard constraints} & \textbf{Norm. / direction} & \textbf{FDE $R$}\\
\midrule\endfirsthead
\multicolumn{7}{@{}c@{}}{\small\tablename~\thetable: Case-level evaluator reference (continued).}\\[2pt]
\toprule\textbf{No.} & \textbf{Case} & \textbf{Fam.} & \textbf{Raw objective} & \textbf{Hard constraints} & \textbf{Norm. / direction} & \textbf{FDE $R$}\\\midrule\endhead
\midrule\multicolumn{7}{r@{}}{\scriptsize continued on next page}\\\endfoot
\bottomrule\endlastfoot
01 & Vehicle-drag prediction & ML & mean relative $C_d$ error ($\mathrm{Cd\_MRE}$) & coverage, unique finite rows; $0.05\leq\mathrm{Cd}\leq1.5$ & \cellstack{\ratioRx}{\scriptsize(min)} & 0.00357527\\
02 & City delivery routing & CO & combined route cost (distance/time/penalties) & all required stops, feasible routes and valid timestamps & \cellstack{\ratioRx}{\scriptsize(min)} & 281497\\
03 & Crop-yield prediction & ML & yearly mean RMSE & required predictions, finite values and complete test coverage & \cellstack{\ratioRx}{\scriptsize(min)} & 3.77281\\
04 & Electricity-price forecasting & ML & weighted MAE/p95/p99/high-price underprediction error & complete horizon, finite values and aligned timestamps & \cellstack{\ratioRx}{\scriptsize(min)} & 0.48004\\
05 & Flight-booking forecasting & ML & seat-booking forecast score & complete horizon, finite nonnegative forecasts & \cellstack{\ratioxR}{\scriptsize(max)} & 0.160653\\
06 & Ion-beam uniformity & ML & mean WAPE-derived uniformity score & all required rows, finite predictions and valid bounds & \cellstack{\ratioxR}{\scriptsize(max)} & 0.64572\\
07 & Lubricant formulation & ML & weighted formulation quality (property/constraint terms) & valid formulation schema, bounds and required composition & \cellstack{\ratioxR}{\scriptsize(max)} & 98.8783\\
08 & Material-property prediction & ML & weighted material-property prediction quality & complete test coverage and finite predictions & \cellstack{\ratioxR}{\scriptsize(max)} & 0.945334\\
09 & Nuclear-site scheduling & CO & weighted construction schedule score & schema, precedence/resource feasibility and horizon limits & \cellstack{\ratioxR}{\scriptsize(max)} & 97.2044\\
10 & Retail stock replenishment & CO & fulfillment/lateness/cost absolute score & complete orders and feasible inventory/transfer records & \cellstack{\ratioxR}{\scriptsize(max)} & 0.997476\\
11 & Retail-zone routing & CO & partition plus routing composite quality & all zones/items covered and routing/partition checks pass & \cellstack{\ratioxR}{\scriptsize(max)} & 0.837337\\
12 & Steel slitting and cutting & CO & cut utilization and cut-count score & valid cut plan, demand coverage and capacity checks & \cellstack{\ratioxR}{\scriptsize(max)} & 1.6645\\
13 & Vessel stowage planning & CO & weighted vessel-stowage objective components & schema, placement, capacity and stability checks & \cellstack{\ratioxR}{\scriptsize(max)} & 0.500878\\
14 & Warehouse stock levels & CO & mean daily policy score minus variability penalty & valid daily policy schema and simulator checks & \cellstack{\ratioxR}{\scriptsize(max)} & 105.697\\
15 & Truck-load planning & CO & $0.60$ coverage $+0.25$ length utilization $+0.15$ balance & all items assigned once and truck/capacity checks & \cellstack{\ratioxR}{\scriptsize(max)} & 1\\
16 & Multi-table APS scheduling & CO & delay/utilization absolute schedule score & all jobs assigned and precedence/resource constraints & \cellstack{\ratioxR}{\scriptsize(max)} & 0.114446\\
17 & Cold-rolling sequencing & CO & coil-sequencing penalty & complete sequence, no duplicates and transition feasibility & \cellstack{\ratioRx}{\scriptsize(min)} & 10718.9\\
18 & Multi-batch scheduling & CO & multi-batch scheduling composite score & schema, assignment, precedence and capacity checks & \cellstack{\ratioxR}{\scriptsize(max)} & 0.998438\\
19 & Multi-line scheduling & CO & total production penalty & complete schedule and line/time feasibility & \cellstack{\ratioRx}{\scriptsize(min)} & 637600\\
20 & Multi-warehouse shipping & CO & lexicographic shipping score with tie-break cost & all orders served and inventory/route constraints & \cellstack{\ratioxR}{\scriptsize(max)} & 0.674404\\
21 & Soybean-meal rescheduling & CO & weighted smoothness/safety/balance rescheduling score & valid schedule and safety/quality checks & \cellstack{\ratioxR}{\scriptsize(max)} & 87.061\\
22 & Wheel-hub scheduling & CO & total weighted delay & two-stage schedule complete and all hard constraints & \cellstack{\ratioRx}{\scriptsize(min)} & 924\\
23 & Auction-picking routes & CO & picking-route distance & all required picks served with valid route & \cellstack{\ratioRx}{\scriptsize(min)} & 237738\\
24 & Quay-crane scheduling & CO & $100{,}000{,}000\,n_{\rm cranes}$ $+$ total move seconds & all hard quay-crane constraints C1--C7 & \cellstack{\ratioRx}{\scriptsize(min)} & 1.1e9\\
25 & Container loading & CO & container stability/loading quality & all items placed once; weight/geometry/stability checks & \cellstack{\ratioxR}{\scriptsize(max)} & 62.0141\\
26 & Three-SKU packing & CO & validated ITEM\_C count/value & valid container schema and packing/weight checks & \cellstack{\ratioxR}{\scriptsize(max)} & 148.006\\
27 & Flight-crew rostering & CO & crew-roster combined score & complete roster and labor/rest/qualification rules & \cellstack{\ratioxR}{\scriptsize(max)} & 4624.98\\
28 & Elevator-call forecasting & ML & $100/(1+\mathrm{MAE})$ minus skew penalty & complete calls, finite nonnegative predictions & \cellstack{\ratioxR}{\scriptsize(max)} & 21.7277\\
29 & Hydropower dispatch & CO & total cascade-hydropower energy & complete dispatch, bounds and cascade constraints & \cellstack{\ratioxR}{\scriptsize(max)} & 160732\\
30 & Syngas composition & ML & mean $R^2$ across syngas targets & complete target coverage and finite predictions & \cellstack{\ratioxR}{\scriptsize(max)} & 0.846809\\
31 & Credit-default scoring & ML & $0.5\,\mathrm{AUC}+0.5\,\mathrm{KS}$ & complete labels/probabilities and valid score range & \cellstack{\ratioxR}{\scriptsize(max)} & 0.865552\\
32 & Grid-load forecasting & ML & MAPE & complete time horizon and finite forecasts & \cellstack{\ratioRx}{\scriptsize(min)} & 0.374465\\
33 & LLM-traffic forecasting & ML & traffic RMSE & complete horizon and finite forecasts & \cellstack{\ratioRx}{\scriptsize(min)} & 232222\\
34 & Jewelry-sales forecasting & ML & $0.5$ quantity score $+0.5$ F1 & complete SKU coverage and valid nonnegative forecasts & \cellstack{\ratioxR}{\scriptsize(max)} & 0.193193\\
35 & Corn-futures forecasting & ML & mean forecasting error & complete horizon and finite forecasts & \cellstack{\ratioRx}{\scriptsize(min)} & 0.294579\\
36 & Fermentation outcomes & ML & fermentation MAE & complete targets and finite predictions & \cellstack{\ratioRx}{\scriptsize(min)} & 0.435077\\
37 & MicroGC pillar design & CO & $0.6s_{\rm nrmse}+0.2s_{\rm press}+0.2s_{\rm perim}$ & valid geometry/field outputs and CFD checks & \cellstack{\ratioxR}{\scriptsize(max)} & 0.62696\\
38 & Display-fab scheduling & CO & $60p_{\rm ach}+30o_{\rm ach}+10q_{\rm score}$ & complete schedule and machine/order constraints & \cellstack{\ratioxR}{\scriptsize(max)} & 98.6208\\
39 & Warehouse forklift scheduling & CO & nested warehouse objective & assignment, stack/forklift and temporal constraints & \cellstack{\ratioRx}{\scriptsize(min)} & 39.4605\\
40 & Frailty-risk grouping & ML & $W_{\rm FIT}R^2+W_{\rm RISK}(c-0.5)$ & complete cohort outputs and valid risk/fit fields & \cellstack{\ratioxR}{\scriptsize(max)} & 0.845504\\
41 & Casting thermal fields & ML & mean thermal-field RMSE & complete field output and finite values & \cellstack{\ratioRx}{\scriptsize(min)} & 0.18952\\
42 & Superalloy strength prediction & ML & superalloy prediction error & complete target coverage and finite values & \cellstack{\ratioRx}{\scriptsize(min)} & 13.3398\\
43 & Coagulant dosing & ML & water-coagulant dosing error & complete horizon and finite values & \cellstack{\ratioRx}{\scriptsize(min)} & 0.21319\\
44 & Bank-branch siting scores & ML & branch-siting score MAE & all branches scored with valid predictions & \cellstack{\ratioRx}{\scriptsize(min)} & 7.12993\\
45 & Data-center rack allocation & CO & rack-allocation total cost & all racks assigned and capacity/power constraints & \cellstack{\ratioRx}{\scriptsize(min)} & 28.7197\\
46 & Routing with 3D loading & CO & $100{,}000\,n_{\rm vehicles}$ $+$ routing distance & packing and route feasibility & \cellstack{\ratioRx}{\scriptsize(min)} & 1.502e6\\
47 & Regional fleet dispatch & CO & regional dispatch total objective & all orders assigned; violations penalized in objective & \cellstack{\ratioRx}{\scriptsize(min)} & 1.452e6\\
48 & Hybrid-energy dispatch & CO & 8,760-hour total income & complete dispatch and operational constraints & \cellstack{\ratioxR}{\scriptsize(max)} & 1.854e8\\
49 & Fuel-tanker routing & CO & multi-compartment tanker distance & all demands served and vehicle/compartment constraints & \cellstack{\ratioRx}{\scriptsize(min)} & 2.827e6\\
\end{longtable}
\endgroup
\paragraph{Clarification points.}\label{app:clarification-qc}
A clarification point is retained only if (i) it cannot be resolved from the
agent-visible request and data, (ii) the customer's answer can change a
feasible decision, the objective, or the evaluator's acceptance judgment, and
(iii) the answer can be stated as a fact, rule, or convention that the
evaluator applies to the artifact.  Implementation preferences, facts visible
in the workspace, and questions about the agent's method are excluded.

\paragraph{Splitting and merging.}
A candidate is split when two answers change different decisions or
evaluator checks, e.g., the error metric, which changes the score, and the
valid value range, which changes whether the artifact is accepted.  Paraphrases and examples of the same decision are merged into
one point.  Table~\ref{tab:clarification-qc-examples} lists boundary cases.

\begin{table}[h]
\centering
\small
\setlength{\tabcolsep}{4pt}
\renewcommand{\arraystretch}{1.12}
\caption{\textbf{Boundary examples for clarification points.}}
\label{tab:clarification-qc-examples}
\begin{tabularx}{\linewidth}{@{}>{\raggedright\arraybackslash}p{0.40\linewidth} >{\raggedright\arraybackslash}X >{\raggedright\arraybackslash}p{0.30\linewidth}@{}}
\toprule
\textbf{Candidate} & \textbf{Decision} & \textbf{Reason}\\
\midrule
``Is error MAE or relative error?'' & Keep as one valid point & Changes the evaluator's scoring convention.\\
``Does same-city supplier precedence override cost minimization?'' & Keep as one valid point & Changes a routing/fulfillment rule and can change feasibility.\\
``Are out-of-range predictions rejected?'' & Keep as one valid point & Changes a hard constraint.\\
``What Python library should I use?'' & Exclude & Implementation choice is not scored.\\
``How many rows are in the visible CSV?'' & Exclude & The fact is already observable to the agent.\\
``Which metric and which value range are required?'' & Split into two points & Metric changes the score; range changes acceptance.\\
``Should same-city orders precede transfers, and should they be grouped?'' & Merge when the evaluator has one precedence rule; split otherwise & The split is determined by independent counterfactual answers, not punctuation.\\
\bottomrule
\end{tabularx}
\end{table}

\paragraph{Registry fields.}
Each point in \texttt{gt.json} has a key and four fields: the requirement
description, issue type (missing, ambiguous, invalidated, or conflicting),
content domain (scoring convention, constraint rule, computation convention,
or data semantics), and rule source (\texttt{null} if none).

\paragraph{Review protocol.}
The FDE responsible for each original delivery annotated the requirements
established in the customer--FDE discussions.  A second review applied the
same criteria against the evaluator and reference artifact, removed
duplicates, and confirmed that each point had a customer answer;
disagreements were resolved against the evaluator code.

\subsection{Metrics and runtime settings}
\label{app:delivery-metrics}

\paragraph{Score selection.}
Up to five extraction routes score each artifact, with remaining routes
skipped once three scores agree after rounding to four decimal places.
Extractions without source provenance are rejected; majority agreement alone
does not prove that every
decision was preserved (Appendix~\ref{app:extraction-audit}).  Without a
majority, the score is unavailable.

\paragraph{Aggregation.}
A run is valid when its artifact satisfies the case's hard constraints
(Table~\ref{tab:case-evaluation-reference}); pass@3 and pass\textasciicircum{}3
count valid repeats.  Zero scores and invalid runs are included in Quality.
Each main configuration contributes three scored runs on each of 49 cases,
with all 147 runs included in DSR (Section~\ref{sec:metrics}). All ablation
runs also have numeric scores. Clarification metrics are never imputed,
and runtime, tool calls,
and cost are not part of any delivery metric.
\paragraph{Question units.}\label{app:process-metrics}
Agent messages are split at question marks into question units; a message
without a question mark counts as one unit.  Turn indices come from the
interaction log.

\paragraph{Coverage judgments.}
Judges also receive annotated trigger examples as matching hints.  Merely
touching a requirement's topic does not count as coverage.

A covered point's matched questions are those cited
by the agreeing judges, and Precision counts distinct matched questions.

\paragraph{Undefined values.}
With no recorded questions, Precision and Ask-F1 are undefined and Recall
is zero.  If every judgment in a run fails, all three metrics are
unavailable; each metric is averaged only over runs where it is defined.
\paragraph{Runtime.}\label{app:runtime}
Each solving run has 43,200 seconds, including interaction, and at most 30
clarification rounds, with no cap on the number of questions or on cumulative
tokens.  OpenHands runs at most 200 agent steps (\texttt{max\_iterations});
these steps are distinct from customer exchanges.  OpenCode passes a per-call
output limit of 131,072 tokens with a 1,000,000-token context.  The
native CLIs use SDK or provider defaults.  OpenCode sets neither temperature
nor reasoning effort, and no run overrides the temperature.
Exact runner, scaffold, and dependency pins are provided with the artifact;
the controls above are the settings relevant to interpreting the experiments.


\begin{table}[p]
\centering
\caption{\textbf{Complete results.} Estimates with 95\% case-bootstrap intervals; bold marks blockwise best values.}
\label{tab:full-configuration-results}\label{tab:full-clarification-results}
{\par\smallskip\noindent{\small\textbf{(a)} Delivery}\par\smallskip
\scriptsize\setlength{\tabcolsep}{3pt}\renewcommand{\arraystretch}{1.05}

\begin{tabularx}{\textwidth}{@{}l *{4}{>{\arraybackslash}X}@{}}
\toprule\rowcolor{DELIVERHeader}\textbf{Agent / model} & \textbf{Quality} & \textbf{DSR} & \textbf{pass@3} & \textbf{pass\textasciicircum{}3}\\\midrule
\rowcolor{DELIVERTotal}
\multicolumn{5}{@{}l}{\emph{General scaffold: \texttt{opencode}}} \\
GPT-6-Astra & \textbf{0.546} [0.448, 0.647] & \textbf{5.4} [0.7, 12.2] & 91.8 [83.7, 98.0] & \textbf{81.6} [71.4, 91.8] \\
GLM-5.3 & 0.541 [0.445, 0.632] & 1.4 [0.0, 3.4] & 91.8 [83.7, 98.0] & 77.6 [65.3, 87.8] \\
Qwen3.8-Max & 0.539 [0.444, 0.629] & 2.7 [0.0, 6.1] & \textbf{95.9} [89.8, 100.0] & 73.5 [61.2, 85.7] \\
Grok-4.6 & 0.522 [0.423, 0.621] & 2.7 [0.7, 5.4] & 87.8 [77.6, 95.9] & 79.6 [67.3, 89.8] \\
DeepSeek-V4-Pro & 0.499 [0.404, 0.593] & 1.4 [0.0, 4.1] & 91.8 [83.7, 98.0] & 71.4 [59.2, 83.7] \\
Kimi-K3 & 0.419 [0.336, 0.503] & 0.0 [0.0, 0.0] & 91.8 [83.7, 98.0] & 53.1 [38.8, 67.3] \\
Gemini-3.1-Pro & 0.387 [0.309, 0.470] & 0.0 [0.0, 0.0] & 89.8 [81.6, 98.0] & 59.2 [46.9, 73.5] \\
\addlinespace[2pt]
\rowcolor{DELIVERTotal}
\multicolumn{5}{@{}l}{\emph{General scaffold: \texttt{openhands}}} \\
GPT-6-Astra & \textbf{0.530} [0.427, 0.624] & 4.8 [0.0, 10.9] & \textbf{93.9} [87.8, 100.0] & 67.3 [53.1, 79.6] \\
GLM-5.3 & 0.527 [0.431, 0.615] & 2.0 [0.0, 5.4] & 91.8 [83.7, 98.0] & 67.3 [55.1, 79.6] \\
Grok-4.6 & 0.516 [0.418, 0.601] & 0.7 [0.0, 2.0] & 91.8 [83.7, 98.0] & \textbf{71.4} [59.2, 83.7] \\
DeepSeek-V4-Pro & 0.510 [0.418, 0.598] & 0.7 [0.0, 2.0] & \textbf{93.9} [85.7, 100.0] & \textbf{71.4} [59.2, 83.7] \\
Qwen3.8-Max & 0.507 [0.421, 0.592] & \textbf{6.1} [1.4, 11.6] & \textbf{93.9} [87.8, 100.0] & 65.3 [53.0, 77.6] \\
Gemini-3.1-Pro & 0.457 [0.360, 0.548] & 0.0 [0.0, 0.0] & 87.8 [77.6, 95.9] & 65.3 [51.0, 77.6] \\
Kimi-K3 & 0.412 [0.331, 0.496] & 0.7 [0.0, 2.0] & 85.7 [75.5, 93.9] & 57.1 [42.9, 71.4] \\
\addlinespace[2pt]
\rowcolor{DELIVERTotal}
\multicolumn{5}{@{}l}{\emph{Native scaffolds}} \\
\mbox{\texttt{codex} + GPT-6-Astra} & \textbf{0.551} [0.448, 0.647] & \textbf{4.1} [0.0, 9.5] & \textbf{93.9} [85.7, 100.0] & \textbf{75.5} [63.3, 87.8] \\
\mbox{\texttt{deepseek-harness} + DeepSeek-V4-Pro} & 0.527 [0.427, 0.619] & 1.4 [0.0, 3.4] & 91.8 [83.7, 98.0] & \textbf{75.5} [63.3, 87.8] \\
\mbox{\texttt{kimi} + Kimi-K3} & 0.420 [0.333, 0.502] & 0.7 [0.0, 2.0] & 89.8 [81.6, 98.0] & 59.2 [44.9, 73.5] \\
\mbox{\texttt{gemini} + Gemini-3.1-Pro} & 0.404 [0.319, 0.494] & 0.0 [0.0, 0.0] & 87.8 [77.6, 95.9] & 57.1 [42.9, 71.4] \\
\bottomrule\end{tabularx}
\par}
\medskip
{\par\smallskip\noindent{\small\textbf{(b)} Clarification}\par\smallskip
\scriptsize\setlength{\tabcolsep}{2.5pt}\renewcommand{\arraystretch}{1.3}

\begin{tabularx}{\textwidth}{@{}l *{6}{>{\arraybackslash}X}@{}}
\toprule\rowcolor{DELIVERHeader}\textbf{Agent / model} & \textbf{Recall} & \textbf{Precision} & \textbf{Ask-F1} & \textbf{KQC$_1$} & \textbf{ATC $\downarrow$} & \textbf{Redund.\ $\downarrow$}\\\midrule
\rowcolor{DELIVERTotal}
\multicolumn{7}{@{}l}{\emph{General scaffold: \texttt{opencode}}} \\
GPT-6-Astra & \cellstack{\textbf{83.6}}{[79.5, 87.6]} & \cellstack{36.5}{[32.1, 41.1]} & \cellstack{\textbf{48.8}}{[44.3, 53.7]} & \cellstack{\textbf{78.1}}{[73.3, 82.6]} & \cellstack{1.07}{[1.04, 1.11]} & \cellstack{15.6}{[13.7, 17.8]} \\
GLM-5.3 & \cellstack{66.8}{[60.7, 72.6]} & \cellstack{\textbf{39.7}}{[35.3, 44.2]} & \cellstack{47.5}{[43.2, 51.8]} & \cellstack{62.8}{[57.2, 68.7]} & \cellstack{1.05}{[1.02, 1.08]} & \cellstack{10.1}{[8.7, 11.8]} \\
Qwen3.8-Max & \cellstack{70.6}{[65.4, 76.0]} & \cellstack{35.4}{[31.5, 39.6]} & \cellstack{45.1}{[40.8, 49.5]} & \cellstack{68.0}{[62.8, 73.3]} & \cellstack{1.03}{[1.01, 1.06]} & \cellstack{16.3}{[14.0, 19.1]} \\
Grok-4.6 & \cellstack{68.2}{[62.3, 74.3]} & \cellstack{38.3}{[34.0, 43.0]} & \cellstack{46.4}{[42.3, 50.6]} & \cellstack{65.2}{[58.9, 71.8]} & \cellstack{1.04}{[1.02, 1.08]} & \cellstack{8.8}{[7.4, 10.5]} \\
DeepSeek-V4-Pro & \cellstack{55.1}{[48.7, 61.9]} & \cellstack{37.9}{[33.3, 42.8]} & \cellstack{42.7}{[38.1, 47.3]} & \cellstack{52.9}{[46.4, 59.9]} & \cellstack{1.04}{[1.01, 1.07]} & \cellstack{8.3}{[7.2, 9.6]} \\
Kimi-K3 & \cellstack{55.0}{[49.3, 60.8]} & \cellstack{37.6}{[33.8, 41.6]} & \cellstack{43.3}{[39.6, 47.3]} & \cellstack{53.6}{[48.3, 59.5]} & \cellstack{\textbf{1.02}}{[1.01, 1.04]} & \cellstack{10.0}{[8.4, 11.7]} \\
Gemini-3.1-Pro & \cellstack{47.1}{[41.3, 52.9]} & \cellstack{39.5}{[34.3, 44.8]} & \cellstack{40.4}{[35.8, 45.4]} & \cellstack{43.0}{[37.4, 48.7]} & \cellstack{1.16}{[1.06, 1.29]} & \cellstack{\textbf{7.5}}{[6.0, 9.0]} \\
\addlinespace[2pt]
\rowcolor{DELIVERTotal}
\multicolumn{7}{@{}l}{\emph{General scaffold: \texttt{openhands}}} \\
GPT-6-Astra & \cellstack{\textbf{84.7}}{[80.5, 88.5]} & \cellstack{36.8}{[32.7, 41.3]} & \cellstack{\textbf{49.5}}{[44.8, 54.0]} & \cellstack{\textbf{80.2}}{[75.2, 85.1]} & \cellstack{1.06}{[1.03, 1.10]} & \cellstack{14.4}{[13.0, 15.8]} \\
GLM-5.3 & \cellstack{67.1}{[61.1, 73.0]} & \cellstack{36.8}{[32.4, 41.1]} & \cellstack{45.5}{[41.1, 49.9]} & \cellstack{64.3}{[58.2, 70.7]} & \cellstack{1.04}{[1.02, 1.06]} & \cellstack{12.6}{[10.7, 14.8]} \\
Grok-4.6 & \cellstack{69.9}{[64.0, 75.9]} & \cellstack{37.0}{[32.7, 41.0]} & \cellstack{45.9}{[41.9, 49.8]} & \cellstack{66.7}{[60.4, 73.1]} & \cellstack{1.05}{[1.02, 1.08]} & \cellstack{11.0}{[9.1, 13.2]} \\
DeepSeek-V4-Pro & \cellstack{65.2}{[58.6, 72.0]} & \cellstack{36.4}{[31.6, 40.9]} & \cellstack{44.9}{[40.1, 49.4]} & \cellstack{64.3}{[57.6, 71.1]} & \cellstack{\textbf{1.01}}{[1.00, 1.02]} & \cellstack{11.0}{[9.6, 12.6]} \\
Qwen3.8-Max & \cellstack{75.4}{[69.8, 80.7]} & \cellstack{30.9}{[27.0, 34.5]} & \cellstack{42.2}{[38.0, 46.5]} & \cellstack{74.0}{[68.1, 79.7]} & \cellstack{1.02}{[1.01, 1.03]} & \cellstack{22.6}{[19.9, 25.6]} \\
Gemini-3.1-Pro & \cellstack{45.5}{[39.0, 52.1]} & \cellstack{\textbf{37.8}}{[32.6, 42.8]} & \cellstack{39.2}{[33.9, 44.5]} & \cellstack{42.2}{[36.2, 48.5]} & \cellstack{1.06}{[1.03, 1.09]} & \cellstack{\textbf{6.3}}{[5.3, 7.4]} \\
Kimi-K3 & \cellstack{57.9}{[52.4, 63.4]} & \cellstack{35.4}{[31.7, 39.1]} & \cellstack{41.3}{[37.8, 44.9]} & \cellstack{57.1}{[51.6, 62.9]} & \cellstack{1.02}{[1.01, 1.04]} & \cellstack{11.2}{[10.0, 12.5]} \\
\addlinespace[2pt]
\rowcolor{DELIVERTotal}
\multicolumn{7}{@{}l}{\emph{Native scaffolds}} \\
\mbox{\texttt{codex} + GPT-6-Astra} & \cellstack{\textbf{84.7}}{[80.8, 88.7]} & \cellstack{37.9}{[33.2, 42.6]} & \cellstack{\textbf{50.1}}{[45.6, 54.6]} & \cellstack{\textbf{79.4}}{[74.9, 83.7]} & \cellstack{1.06}{[1.03, 1.09]} & \cellstack{13.9}{[12.2, 15.9]} \\
\mbox{\texttt{deepseek-harness} + DeepSeek-V4-Pro} & \cellstack{61.8}{[55.3, 68.1]} & \cellstack{\textbf{39.3}}{[34.8, 43.8]} & \cellstack{45.9}{[41.0, 50.2]} & \cellstack{59.9}{[53.5, 66.4]} & \cellstack{1.03}{[1.01, 1.06]} & \cellstack{9.6}{[8.1, 11.4]} \\
\mbox{\texttt{kimi} + Kimi-K3} & \cellstack{55.7}{[49.3, 62.3]} & \cellstack{36.7}{[32.4, 41.1]} & \cellstack{42.9}{[38.3, 47.3]} & \cellstack{54.2}{[47.9, 60.8]} & \cellstack{1.03}{[1.01, 1.05]} & \cellstack{9.8}{[8.3, 11.6]} \\
\mbox{\texttt{gemini} + Gemini-3.1-Pro} & \cellstack{44.9}{[39.3, 50.6]} & \cellstack{38.8}{[33.6, 44.1]} & \cellstack{39.4}{[34.5, 44.2]} & \cellstack{44.2}{[38.4, 50.3]} & \cellstack{\textbf{1.01}}{[1.00, 1.03]} & \cellstack{\textbf{6.0}}{[5.1, 6.9]} \\
\bottomrule\end{tabularx}
\par}
\end{table}

\subsection{Full results}
\label{app:full-results}

Entries are means with 95\% case-bootstrap intervals (2,000 resamples, seed
12345), following the bootstrap framework of Efron~\citep{efron1992bootstrap}.

\paragraph{Delivery metrics.}

Table~\ref{tab:full-configuration-results}(a) gives delivery results with
intervals.  Quality is case-macro and DSR is run-level; pass@3 and
pass\textasciicircum{}3 are case-level (Section~\ref{sec:metrics}).  Every
pair of unpaired Quality intervals overlaps, because the intervals carry the
variance between cases.  Paired by case, the systems form two groups.  The 12
systems built on GPT-6-Astra, GLM-5.3, Qwen3.8-Max, Grok-4.6, and
DeepSeek-V4-Pro score 0.499--0.551; the three Kimi-K3 systems and
Gemini-3.1-Pro under \texttt{opencode} and \texttt{gemini} score
0.387--0.420.  All 60 differences between the groups exclude zero under a
paired case bootstrap, and none of the 76 differences within a group does.
All differences above 0.08 are resolved and none of 0.04 or less, so 49 cases
separate models but not scaffolds of the same model.  The exception is
\texttt{openhands} + Gemini-3.1-Pro (0.457), which exceeds the other two
Gemini-3.1-Pro systems by 0.053 and 0.070 (both resolved).  DSR does not rank
systems: 15 of the 18 DSR intervals include zero.

\paragraph{Clarification metrics.}
Clarification metrics are run-level (Section~\ref{sec:clarification-metrics};
Table~\ref{tab:full-clarification-results}(b)). Across the 18 systems,
Spearman $\rho$ between Recall and Quality is 0.83; after subtracting each model's mean over
its scaffolds, it is $-0.31$ ($p=0.22$).  Moving a model from
\texttt{opencode} to \texttt{openhands} raises Recall for six of seven models
(DeepSeek-V4-Pro: 55.1\% to 65.2\%) and Redund.\ for five (Qwen3.8-Max: 16.3
to 22.6), lowers Precision for six, and lowers Quality for five, by at most
0.032. Scaffold changes that increase coverage do not consistently improve
Quality; the policy ablation also shows mixed Quality changes
(Appendix~\ref{app:clarification-policy}).

\paragraph{Compute.}
Provider cost was available for 15 of 18 configurations and showed no
detectable association with Quality; detailed accounting is included with the
artifact rather than repeated here.
\subsection{Clarification types}
\label{app:clarification-types}

Each clarification point in \texttt{gt.json} carries an issue type and a
content domain (Table~\ref{tab:clarification-type-composition}).  These labels
are hidden from agents and judges and are not used by the primary scorer.

\begin{table}[h]
\centering
\small
\setlength{\tabcolsep}{4pt}
\renewcommand{\arraystretch}{1.08}
\caption{\textbf{Composition of the clarification-point registry.} 266 points in 49 cases.}
\label{tab:clarification-type-composition}
\begin{tabularx}{\linewidth}{@{}>{\raggedright\arraybackslash}X l >{\centering\arraybackslash}X >{\centering\arraybackslash}X >{\centering\arraybackslash}X@{}}
\toprule
\rowcolor{DELIVERHeader}
\textbf{Annotation family} & \textbf{Type} & \textbf{Points} &
\textbf{Share} & \textbf{Cases} \\
\midrule
\multirow{4}{*}{Issue type}
& Missing & 207 & 77.8\% & 48 \\
& Ambiguous & 45 & 16.9\% & 30 \\
& Invalidated & 9 & 3.4\% & 8 \\
& Conflicting & 5 & 1.9\% & 4 \\
\addlinespace[2pt]
\multirow{4}{*}{Content domain}
& Scoring convention & 78 & 29.3\% & 39 \\
& Constraint rule & 65 & 24.4\% & 33 \\
& Computation convention & 62 & 23.3\% & 27 \\
& Data semantics & 61 & 22.9\% & 35 \\
\bottomrule
\end{tabularx}
\end{table}

Rule-source labels are too sparse (246 unlabeled) to analyze.  Type-specific Recall is the share of a
case's points of that type that are covered, averaged over repeats and then
over cases containing the type, and correlated with Quality across the 18
configurations (Table~\ref{tab:clarification-type-overall}(a)).

\begin{table}[h]
\centering
\caption{\textbf{Issue types and delivery.} Spearman $\rho$ with case-bootstrap 95\% intervals.}
\label{tab:clarification-type-overall}\label{tab:clarification-type-case}
{\par\smallskip\noindent{\small\textbf{(a)} Type-specific Recall and Quality across 18 configurations; Holm correction over four tests}\par\smallskip
\small
\setlength{\tabcolsep}{4pt}
\renewcommand{\arraystretch}{1.08}

\begin{tabularx}{\linewidth}{@{}>{\raggedright\arraybackslash}X *{4}{>{\arraybackslash}X}@{}}
\toprule
\rowcolor{DELIVERHeader}
\textbf{Issue type} & \textbf{Mean Recall} & \textbf{Quality $\rho$} &
\textbf{95\% CI} & \textbf{Holm $p$} \\
\midrule
Missing & 66.3\% & 0.808 & [0.467, 0.882] & $<0.001$ \\
Ambiguous & 60.5\% & 0.801 & [0.423, 0.884] & $<0.001$ \\
Invalidated & 60.5\% & 0.513 & [0.004, 0.707] & 0.029 \\
Conflicting & 77.5\% & 0.641 & [0.101, 0.705] & 0.008 \\
\bottomrule
\end{tabularx}
\par}
\medskip
{\par\smallskip\noindent{\small\textbf{(b)} Issue-type share and delivery across 49 cases}\par\smallskip
\small
\setlength{\tabcolsep}{3.5pt}
\renewcommand{\arraystretch}{1.08}

\begin{tabularx}{\linewidth}{@{}>{\raggedright\arraybackslash}X >{\arraybackslash}X *{2}{>{\arraybackslash}X}@{}}
\toprule
\rowcolor{DELIVERHeader}
\textbf{Issue type} & \textbf{Cases with type} & \textbf{Quality $\rho$ [95\% CI]} &
\textbf{DSR $\rho$ [95\% CI]} \\
\midrule
Missing & 48 & \cellstack{0.032}{\scriptsize[-0.253, 0.341]} &
\cellstack{$-0.092$}{\scriptsize[-0.322, 0.161]} \\
Ambiguous & 30 & \cellstack{$-0.081$}{\scriptsize[-0.358, 0.209]} &
\cellstack{0.018}{\scriptsize[-0.212, 0.260]} \\
Invalidated & 8 & \cellstack{0.010}{\scriptsize[-0.274, 0.283]} &
\cellstack{0.414}{\scriptsize[0.021, 0.747]} \\
Conflicting & 4 & \cellstack{$-0.102$}{\scriptsize[-0.357, 0.216]} &
\cellstack{0.085}{\scriptsize[-0.159, 0.457]} \\
\bottomrule
\end{tabularx}
\par}
\end{table}

Table~\ref{tab:type-recall-by-config} breaks these Recall values down by
configuration.

\begin{table}[h]
\centering
\fontsize{8}{9}\selectfont
\setlength{\tabcolsep}{3pt}
\renewcommand{\arraystretch}{1.02}
\caption{\textbf{Type-specific Recall (\%) by configuration.} Bold marks blockwise maxima.}
\label{tab:type-recall-by-config}
\begin{tabularx}{\linewidth}{@{}l *{5}{>{\centering\arraybackslash}X}@{}}
\toprule
\rowcolor{DELIVERHeader}
\textbf{Agent / model} & \textbf{Overall} & \textbf{Missing} & \textbf{Ambiguous} & \textbf{Invalidated} & \textbf{Conflicting} \\
\midrule
\rowcolor{DELIVERTotal}
\multicolumn{6}{@{}l}{\emph{General scaffold: \texttt{opencode}}} \\
GPT-6-Astra & \textbf{83.64} & \textbf{87.8} & \textbf{75.7} & 64.6 & \textbf{95.8} \\
GLM-5.3 & 66.83 & 69.0 & 63.1 & 56.2 & 75.0 \\
Qwen3.8-Max & 70.56 & 70.7 & 67.8 & \textbf{81.2} & 83.3 \\
Grok-4.6 & 68.19 & 71.2 & 66.5 & 64.6 & 70.8 \\
DeepSeek-V4-Pro & 55.08 & 58.0 & 46.5 & 52.1 & 70.8 \\
Kimi-K3 & 55.03 & 58.1 & 48.3 & 62.5 & 75.0 \\
Gemini-3.1-Pro & 47.10 & 49.9 & 43.9 & 41.7 & 54.2 \\
\addlinespace[2pt]
\rowcolor{DELIVERTotal}
\multicolumn{6}{@{}l}{\emph{General scaffold: \texttt{openhands}}} \\
GPT-6-Astra & \textbf{84.67} & \textbf{87.2} & \textbf{78.3} & 75.0 & 87.5 \\
GLM-5.3 & 67.12 & 68.9 & 58.9 & 62.5 & 75.0 \\
Grok-4.6 & 69.91 & 71.7 & 64.4 & 72.9 & 83.3 \\
DeepSeek-V4-Pro & 65.24 & 65.6 & 63.7 & 60.4 & 87.5 \\
Qwen3.8-Max & 75.37 & 77.2 & 68.0 & \textbf{85.4} & \textbf{100.0} \\
Gemini-3.1-Pro & 45.45 & 45.7 & 46.5 & 33.3 & 50.0 \\
Kimi-K3 & 57.94 & 59.9 & 52.0 & 68.8 & 62.5 \\
\addlinespace[2pt]
\rowcolor{DELIVERTotal}
\multicolumn{6}{@{}l}{\emph{Native scaffolds}} \\
\mbox{\texttt{codex} + GPT-6-Astra} & \textbf{84.70} & \textbf{87.9} & \textbf{80.0} & \textbf{70.8} & \textbf{100.0} \\
\mbox{\texttt{deepseek-harness} + DeepSeek-V4-Pro} & 61.79 & 62.3 & 62.4 & 56.2 & 83.3 \\
\mbox{\texttt{kimi} + Kimi-K3} & 55.69 & 56.9 & 55.7 & 52.1 & 83.3 \\
\mbox{\texttt{gemini} + Gemini-3.1-Pro} & 44.94 & 45.8 & 46.5 & 29.2 & 58.3 \\
\bottomrule
\end{tabularx}
\end{table}

The ranking of systems is stable across issue types.  Missing and Ambiguous
Recall have Spearman $\rho=0.95$ over the 18 configurations, and GPT-6-Astra
has the highest and Gemini-3.1-Pro the lowest of both in every scaffold
block.  Invalidated points are the exception: Qwen3.8-Max ranks first on them
in both general scaffolds (81.2\% and 85.4\%) although its Missing Recall is
10--17 points below GPT-6-Astra; the Invalidated and Conflicting columns rest
on 8 and 4 cases.  Within a case, the issue type does not decide which points
are covered.  In the 29 cases with both Missing and Ambiguous points, 16 of 18
systems cover Ambiguous points less often (mean gap 7.2 points), but the gap
reverses in 12 of the 29 cases (95\% case-bootstrap CI $[-3.8, 17.7]$).

The content domain does.  Pooled over runs, agents cover 72.6\% of
constraint-rule points, 64.8\% of computation conventions, 56.6\% of scoring
conventions, and 56.6\% of data-semantics points, and constraint rules have
the highest Recall for 16 of 18 systems.  Within the same case, constraint
rules are covered 14.6 points more often than data semantics (22 cases; CI
$[3.9, 25.0]$) and 15.9 points more than computation conventions (22 cases;
$[2.8, 28.9]$); the gap to scoring conventions ($+12.5$, 26 cases) has an
interval that includes zero.  Agents ask about stated business rules more
than about how fields are read and how quantities are computed.  Covering a
constraint rule also has the largest association with delivery: within the
same system and case, runs that cover a constraint-rule point score 0.059
higher than runs that do not (292 system--case--point cells with mixed
coverage; Wilcoxon $p=0.006$), against 0.027--0.035 for the other domains.

Across systems, coverage of every domain tracks Quality equally (Spearman
$\rho=0.81$--$0.83$), and the two groups of Appendix~\ref{app:full-results}
differ by 16.2--20.7 points in every domain.  Table~\ref{tab:question-mix}
gives the domain mix of each model's matched questions; a question matched to
points of several domains is split equally among them.  GPT-6-Astra is the
only model that directs more of its matched questions to scoring conventions
(28.8\%) than to constraint rules (26.7\%); DeepSeek-V4-Pro, Gemini-3.1-Pro,
and Grok-4.6 direct 20.1--21.7\% to scoring conventions and 31.5--33.0\% to
constraint rules.  Over the 18 systems, the scoring share correlates with
Quality ($\rho=0.61$) and the constraint share with lower Quality
($\rho=-0.60$; Holm $p=0.028$ for both over four domains).  Neither
association is robust: the case-bootstrap 95\% intervals include zero
($[-0.07, 0.64]$ and $[-0.69, 0.08]$), and without the three GPT-6-Astra
systems the correlations fall to 0.38 ($p=0.16$) and $-0.32$ ($p=0.24$).

\begin{table}[h]
\centering
\small
\setlength{\tabcolsep}{4pt}
\renewcommand{\arraystretch}{1.08}
\caption{\textbf{Domain mix of matched questions by model (\%).} Quality is the mean over the model's systems.}
\label{tab:question-mix}
\begin{tabularx}{\linewidth}{@{}l *{5}{>{\centering\arraybackslash}X}@{}}
\toprule
\rowcolor{DELIVERHeader}
\textbf{Model} & \textbf{Constraint} & \textbf{Computation} & \textbf{Scoring} & \textbf{Data} & \textbf{Quality} \\
\midrule
GPT-6-Astra & 26.7 & 25.4 & 28.8 & 19.1 & 0.542 \\
GLM-5.3 & 30.7 & 24.5 & 26.6 & 18.3 & 0.534 \\
Qwen3.8-Max & 28.1 & 27.4 & 26.3 & 18.2 & 0.523 \\
Grok-4.6 & 31.5 & 29.7 & 21.7 & 17.2 & 0.519 \\
DeepSeek-V4-Pro & 33.0 & 27.9 & 20.1 & 19.0 & 0.512 \\
Kimi-K3 & 31.1 & 26.1 & 24.6 & 18.2 & 0.417 \\
Gemini-3.1-Pro & 33.0 & 29.6 & 20.4 & 17.0 & 0.416 \\
\bottomrule
\end{tabularx}
\end{table}

Table~\ref{tab:clarification-type-case}(b) instead correlates each case's share
of points of a type with its delivery, averaged over configurations, across
all 49 cases.

No Quality interval excludes zero, so the type mix of a case does not
predict its difficulty.  The positive Invalidated--DSR association rests on
eight cases.
\subsection{Ablations}
\label{app:ablation-audit}

\paragraph{Scope.}
The three-condition ablation uses a 20-case subset.  The policy ablation uses
the five cases in that subset with the most clarification points; exact case
IDs and registry counts are provided with the artifact.
\paragraph{Information conditions.}\label{app:condition-statistics}
Table~\ref{tab:condition-intervals} expands Table~\ref{tab:condition-effects}.
Each estimate averages 20 paired case differences.  Intervals are 95\%
case-bootstrap intervals (2,000 resamples, seed 12345); $^{*}$ marks
two-sided Wilcoxon signed-rank tests that remain significant after Holm
correction over the 24 comparisons.

\begin{table}[h]
\centering
\fontsize{8}{9.5}\selectfont
\setlength{\tabcolsep}{3pt}
\renewcommand{\arraystretch}{1.2}
\caption{\textbf{Paired Quality differences between information conditions.} $^{*}$: Holm-adjusted $p<0.05$.}
\label{tab:condition-intervals}
\begin{tabularx}{\linewidth}{@{}l *{3}{>{\centering\arraybackslash}X}@{}}
\toprule
\rowcolor{DELIVERHeader}
\textbf{Agent / model} & $\boldsymbol{\Delta_{\rm req}}$ &
$\boldsymbol{\Delta_{\rm remain}}$ & $\boldsymbol{\Delta_{\rm total}}$ \\
\midrule
\texttt{opencode} + GPT-6-Astra & \shortstack{+0.3997$^{*}$\\{[0.2284, 0.5983]}} & \shortstack{+0.1662\\{[0.0369, 0.3167]}} & \shortstack{+0.5658$^{*}$\\{[0.3941, 0.7367]}} \\
\texttt{opencode} + DeepSeek-V4-Pro & \shortstack{+0.3640$^{*}$\\{[0.2367, 0.5046]}} & \shortstack{+0.1758$^{*}$\\{[0.0893, 0.2733]}} & \shortstack{+0.5398$^{*}$\\{[0.4026, 0.6741]}} \\
\texttt{opencode} + Gemini-3.1-Pro & \shortstack{+0.1694$^{*}$\\{[0.0824, 0.2684]}} & \shortstack{+0.1974$^{*}$\\{[0.0810, 0.3276]}} & \shortstack{+0.3668$^{*}$\\{[0.2266, 0.5130]}} \\
\texttt{opencode} + Kimi-K3 & \shortstack{+0.2450$^{*}$\\{[0.1326, 0.3595]}} & \shortstack{+0.0794\\{[-0.0125, 0.1740]}} & \shortstack{+0.3244$^{*}$\\{[0.2137, 0.4438]}} \\
\texttt{codex} + GPT-6-Astra & \shortstack{+0.4000$^{*}$\\{[0.2211, 0.5751]}} & \shortstack{+0.1966$^{*}$\\{[0.0853, 0.3228]}} & \shortstack{+0.5966$^{*}$\\{[0.4348, 0.7473]}} \\
\texttt{deepseek-harness} + DeepSeek-V4-Pro & \shortstack{+0.3644$^{*}$\\{[0.2344, 0.5002]}} & \shortstack{+0.1697$^{*}$\\{[0.0689, 0.2949]}} & \shortstack{+0.5341$^{*}$\\{[0.3960, 0.6677]}} \\
\texttt{gemini} + Gemini-3.1-Pro & \shortstack{+0.2532$^{*}$\\{[0.1422, 0.3757]}} & \shortstack{+0.1420\\{[0.0189, 0.2758]}} & \shortstack{+0.3952$^{*}$\\{[0.2680, 0.5216]}} \\
\texttt{kimi} + Kimi-K3 & \shortstack{+0.2195$^{*}$\\{[0.1044, 0.3488]}} & \shortstack{+0.3490$^{*}$\\{[0.2226, 0.4809]}} & \shortstack{+0.5685$^{*}$\\{[0.4156, 0.7132]}} \\
\bottomrule
\end{tabularx}
\end{table}

$\Delta_{\rm req}$ and $\Delta_{\rm total}$ are significant for all eight
systems; $\Delta_{\rm remain}$ is significant for five.  What interaction
recovers depends on how many requirements the system asks about; what remains
does not.  Over the eight systems, $\Delta_{\rm req}$ rises with Recall on the
same 20 cases (Spearman $\rho=0.86$, $p=0.007$), while $\Delta_{\rm remain}$
has $\rho=0.00$.  The two Kimi-K3 systems show the difference: with Recall of
60.4\% and 60.7\% and the same \texttt{Interact-Req} Quality (0.369), the
native \texttt{kimi} CLI reaches 0.718 under \texttt{Full} and
\texttt{opencode} 0.449, so $\Delta_{\rm remain}$ is $+0.349$ and $+0.079$.
\paragraph{Clarification policy.}\label{app:clarification-policy}
Both policies are run on the five policy cases with the eight ablation
configurations (15 runs per configuration and policy).  \texttt{Interact}
lets the agent decide whether to ask; \texttt{Interact-Req} requires
clarification before implementation.  \#Q counts question units, and ask
rate is the share of runs with at least one question.

\begin{table}[h]
\centering
\caption{\textbf{Optional versus required clarification} on five cases.}
\label{tab:clarification-policy}\label{tab:policy-effects}
{\par\smallskip\noindent{\small\textbf{(a)} Metrics by condition}\par\smallskip
\fontsize{8}{9.5}\selectfont
\setlength{\tabcolsep}{3.5pt}
\renewcommand{\arraystretch}{1.08}

\begin{tabularx}{\linewidth}{@{}l *{6}{>{\arraybackslash}X}@{}}
\toprule
\rowcolor{DELIVERHeader}
\textbf{Condition} & \textbf{Quality} &
\textbf{Ask rate} & $\boldsymbol{\#Q}$ & \textbf{Recall} & \textbf{Precision} &
\textbf{Ask-F1} \\
\midrule
\multicolumn{7}{@{}l}{\emph{\texttt{opencode} + GPT-6-Astra}} \\
\texttt{Interact} & 0.3262 & 100.00\% & 15.00 & 63.61\% & 47.72\% & 53.84\% \\
\texttt{Interact-Req} & 0.3830 & 100.00\% & 36.27 & 85.24\% & 49.04\% & 61.19\% \\
\addlinespace[2pt]
\multicolumn{7}{@{}l}{\emph{\texttt{opencode} + DeepSeek-V4-Pro}} \\
\texttt{Interact} & 0.4061 & 100.00\% & 10.20 & 37.18\% & 56.92\% & 43.73\% \\
\texttt{Interact-Req} & 0.5552 & 100.00\% & 21.13 & 49.76\% & 53.51\% & 50.62\% \\
\addlinespace[2pt]
\multicolumn{7}{@{}l}{\emph{\texttt{opencode} + Gemini-3.1-Pro}} \\
\texttt{Interact} & 0.3304 & 93.33\% & 10.07 & 39.45\% & 53.15\% & 46.42\% \\
\texttt{Interact-Req} & 0.2742 & 100.00\% & 14.47 & 38.53\% & 47.24\% & 41.01\% \\
\addlinespace[2pt]
\multicolumn{7}{@{}l}{\emph{\texttt{opencode} + Kimi-K3}} \\
\texttt{Interact} & 0.0992 & 80.00\% & 8.00 & 23.67\% & 46.33\% & 34.53\% \\
\texttt{Interact-Req} & 0.2955 & 100.00\% & 24.27 & 52.59\% & 43.83\% & 45.28\% \\
\addlinespace[2pt]
\multicolumn{7}{@{}l}{\emph{\texttt{codex} + GPT-6-Astra}} \\
\texttt{Interact} & 0.4612 & 100.00\% & 16.27 & 69.97\% & 51.47\% & 58.61\% \\
\texttt{Interact-Req} & 0.3911 & 100.00\% & 31.27 & 81.03\% & 48.50\% & 59.43\% \\
\addlinespace[2pt]
\multicolumn{7}{@{}l}{\emph{\texttt{deepseek-harness} + DeepSeek-V4-Pro}} \\
\texttt{Interact} & 0.4269 & 80.00\% & 8.33 & 33.09\% & 57.05\% & 46.32\% \\
\texttt{Interact-Req} & 0.5579 & 100.00\% & 18.60 & 53.26\% & 55.03\% & 53.50\% \\
\addlinespace[2pt]
\multicolumn{7}{@{}l}{\emph{\texttt{gemini} + Gemini-3.1-Pro}} \\
\texttt{Interact} & 0.2536 & 93.33\% & 8.00 & 30.73\% & 47.06\% & 38.12\% \\
\texttt{Interact-Req} & 0.3424 & 100.00\% & 13.07 & 41.19\% & 51.12\% & 44.28\% \\
\addlinespace[2pt]
\multicolumn{7}{@{}l}{\emph{\texttt{kimi} + Kimi-K3}} \\
\texttt{Interact} & 0.3603 & 100.00\% & 8.07 & 38.17\% & 56.13\% & 43.33\% \\
\texttt{Interact-Req} & 0.2351 & 86.67\% & 19.47 & 41.54\% & 44.32\% & 43.57\% \\
\addlinespace[2pt]
\bottomrule
\end{tabularx}
\par}
\medskip
{\par\smallskip\noindent{\small\textbf{(b)} Quality change from required clarification, with Holm-adjusted $p$-values}\par\smallskip
\fontsize{8}{9.5}\selectfont
\setlength{\tabcolsep}{4pt}
\renewcommand{\arraystretch}{1.1}

\begin{tabularx}{\linewidth}{@{}l *{3}{>{\arraybackslash}X}@{}}
\toprule
\rowcolor{DELIVERHeader}
\textbf{Agent / model} & \textbf{Req$-$Interact} & \textbf{95\% CI} & \textbf{Holm $p$} \\
\midrule
\texttt{opencode} + GPT-6-Astra & +0.0569 & [-0.0727, 0.2129] & 1.000 \\
\texttt{opencode} + DeepSeek-V4-Pro & +0.1490 & [0.0535, 0.2467] & 0.875 \\
\texttt{opencode} + Gemini-3.1-Pro & -0.0563 & [-0.1866, 0.0383] & 1.000 \\
\texttt{opencode} + Kimi-K3 & +0.1963 & [0.0882, 0.3217] & 0.500 \\
\texttt{codex} + GPT-6-Astra & -0.0701 & [-0.2065, 0.0440] & 1.000 \\
\texttt{deepseek-harness} + DeepSeek-V4-Pro & +0.1310 & [-0.1608, 0.4518] & 1.000 \\
\texttt{gemini} + Gemini-3.1-Pro & +0.0888 & [-0.1103, 0.3339] & 1.000 \\
\texttt{kimi} + Kimi-K3 & -0.1252 & [-0.5651, 0.2291] & 1.000 \\
\bottomrule
\end{tabularx}
\par}
\end{table}

Without the requirement, agents still ask in 80--100\% of runs, but they ask
fewer questions: requiring clarification multiplies \#Q by 1.4--3.0 in all
eight systems.  The added questions are worth less than the ones agents ask
on their own.  Measured as the change in matched questions over the change in
\#Q, they match a covered requirement at 13.5--45.8\%, below the
\texttt{Interact} Precision of the same system in all eight.  Recall rises in
seven systems, but by at most 2.1 points per added question.  Agents ask
their most useful questions first.  Quality changes are mixed
(Table~\ref{tab:policy-effects}(b)): no Holm-adjusted $p$ is below 0.5.

\paragraph{Oracle answer coverage.}\label{app:information-ablation}
We selected four cases (09, 11, 17, and 38) and ran \texttt{codex} +
GPT-6-Astra without a clarification channel.  We evaluate five target
coverage levels, $r\in\{0,0.25,0.5,0.75,1\}$, using nested subsets of the
registered answers and three repeats per case and level.  The 0\% level is
run separately from \texttt{Hidden}; the 100\% level supplies the registered
answers, not the complete information file used by \texttt{Full}.
Table~\ref{tab:info-ablation} gives the values plotted in
Figure~\ref{fig:oracle-coverage}; exact subset construction is provided with
the artifact.

\begin{table}[h]
\centering
\fontsize{8}{9.5}\selectfont
\setlength{\tabcolsep}{3pt}
\renewcommand{\arraystretch}{1.15}
\caption{\textbf{Oracle Quality by answer coverage.} Mean $\pm$ SD over three repeats.}
\label{tab:info-ablation}
\begin{tabularx}{\linewidth}{@{}l *{5}{>{\centering\arraybackslash}X}@{}}
\toprule
\rowcolor{DELIVERHeader}
\textbf{Case} & \textbf{0\%} & \textbf{25\%} & \textbf{50\%} & \textbf{75\%} & \textbf{100\%} \\
\midrule
09 Nuclear-site scheduling & 0.0000 $\pm$ 0.0000 & 0.0000 $\pm$ 0.0000 & 0.3272 $\pm$ 0.5667 & 0.3206 $\pm$ 0.5553 & 0.9770 $\pm$ 0.0109 \\
11 Retail-zone routing & 0.0000 $\pm$ 0.0000 & 0.4661 $\pm$ 0.4037 & 0.7442 $\pm$ 0.0629 & 0.7935 $\pm$ 0.0670 & 0.8558 $\pm$ 0.0582 \\
17 Cold-rolling sequencing & 0.2347 $\pm$ 0.1189 & 0.1637 $\pm$ 0.0399 & 0.3357 $\pm$ 0.1595 & 0.4260 $\pm$ 0.1418 & 0.6782 $\pm$ 0.2787 \\
38 Display-fab scheduling & 0.0000 $\pm$ 0.0000 & 0.0000 $\pm$ 0.0000 & 0.0000 $\pm$ 0.0000 & 0.3190 $\pm$ 0.5525 & 0.9650 $\pm$ 0.0093 \\
\midrule
Four-case mean & 0.0587 & 0.1575 & 0.3518 & 0.4648 & 0.8690 \\
\bottomrule
\end{tabularx}
\end{table}

Answer coverage improves Quality overall, but the case-level trends are
non-monotonic.  Specific answers matter more than the number of answers, which
also explains the high variance at intermediate coverage levels.  The
registered-answer endpoint differs from \texttt{Full}, and coverage of a
blocking point is necessary but not sufficient for correct implementation;
the table therefore provides an information-sensitivity analysis rather than
a claim that more answers always improve delivery.
\subsection{Zero scores and a case study}
\label{app:zero-score-analysis}

Of the 2,646 main runs, 548 score zero.  Of these, 509
(92.9\%) are invalid and 39 (7.1\%) are valid but earn no Quality; all
have a complete artifact and pipeline record.
Table~\ref{tab:zero-score-signatures} groups the recorded \texttt{error\_info}
into mutually exclusive signatures (precedence: constraint, fatal, schema,
execution).  Constraint or feasibility violations dominate (430, 78.5\%), and
61 runs (11.1\%) carry no explicit message.

\begin{table}[h]
\centering
\small
\setlength{\tabcolsep}{4pt}
\renewcommand{\arraystretch}{1.1}
\caption{\textbf{Evaluator signatures of 548 zero-score runs.}}
\label{tab:zero-score-signatures}
\begin{tabularx}{\linewidth}{@{}>{\raggedright\arraybackslash}p{0.28\linewidth} >{\centering\arraybackslash}p{0.05\linewidth} >{\centering\arraybackslash}p{0.06\linewidth} >{\raggedright\arraybackslash}X@{}}
\toprule
\rowcolor{DELIVERHeader}
\textbf{Recorded signature} & \textbf{Runs} & \textbf{Share} & \textbf{Interpretation} \\
\midrule
Constraint/feasibility violation & 430 & 78.5\% & Case constraints or feasibility checks are explicitly violated. \\
Fatal case-specific check & 29 & 5.3\% & A case-defined fatal condition is recorded, such as incomplete coverage or an out-of-horizon event. \\
Schema/output validation error & 8 & 1.5\% & The submitted representation fails a required schema or output check. \\
Evaluator/scoring execution exception & 4 & 0.7\% & The evaluator records an exception or cannot form the required score. \\
Quality/threshold warning & 16 & 2.9\% & A quality or threshold warning accompanies zero Quality for a valid artifact. \\
No explicit evaluator message & 61 & 11.1\% & The artifact is scored zero but the retained record does not identify a more specific reason. \\
\bottomrule
\end{tabularx}
\end{table}

Zero scores belong to cases more than to systems.  They occur in 33 of the
49 cases, and 7 cases account for half of them.  The five cases with the most
zeros (21, 10, 13, 18, and 38) contribute 233 (42.5\%); in each, 17 or 18
of the 18 systems score zero at least once, and in case 21, 17 systems score
zero in all three repeats.  Within a system and case, clarification separates
failing from passing repeats only slightly: in the 202 cells with both zero
and nonzero repeats, zero-score repeats have Recall 58.6\% against 63.6\%
(lower in 89 cells, higher in 72; Wilcoxon $p=0.021$) and ask as many
questions (22.8 vs.\ 23.2).  Seventy-two zeros (13.1\%) reached the runtime
limit but still produced a scored artifact; 54 of them also record a
constraint violation, so a timeout does not by itself explain the zero.
\paragraph{Case study: case 10.}\label{app:case-study-019}
We trace one run of case 10 (retail replenishment; 17 days, 73 stores,
3,055 items; \texttt{opencode} + GPT-6-Astra, \texttt{Interact-Req}, first
repeat); Table~\ref{tab:case-study-019-chain}(a) gives the evidence chain.

\begin{table}[h]
\centering
\caption{\textbf{Case-10 run.}}
\label{tab:case-study-019-chain}\label{tab:case-study-019-item}
{\par\smallskip\noindent{\small\textbf{(a)} Evidence chain}\par\smallskip
\small
\setlength{\tabcolsep}{4pt}
\renewcommand{\arraystretch}{1.12}

\begin{tabularx}{\linewidth}{@{}>{\raggedright\arraybackslash}p{0.19\linewidth} X@{}}
\toprule
\textbf{Stage} & \textbf{Recorded evidence} \\
\midrule
Customer request & JSON replenishment plan; internal store/warehouse transfers
must be exhausted before supplier purchase, and same-city stock cannot be
skipped; stockout avoidance takes priority over timeliness and cost. \\
Agent questions & 27 messages (47 scored question units) over three rounds.
The agent explicitly asked
about source precedence (``same-city stores and local warehouse $\rightarrow$
cross-city sources $\rightarrow$ supplier'') and about how shortages and lead
times should be computed. \\
Key rule and coverage & The complete business information specifies
receive arrivals $\rightarrow$ settle sales $\rightarrow$ dispatch transfers
and purchases.  The registered point ``daily inventory transition order''
requires this exact order.  The stored majority label
is \texttt{covered=false}, with 0.60 judge agreement (2 covered and 3
uncovered votes).  No matched question was recorded for this blocker. \\
Submitted plan & The plan metadata nevertheless declares the same three-step
order.  It contains 15,987 instructions (15,515 transfers and 472 purchases),
and the agent's validator reports \texttt{pass}. \\
Executable evaluation & All five extractor/evaluator routes mark the plan
invalid and return Quality $=0.0$.  The evaluator records 347
constraint violations, including eight same-city exhaustion violations for
\texttt{ITEM\_000092}. \\
Reference & The customer-accepted reference has value 0.9974756.  For the
eight affected stores, it contains no item-092 supplier purchase and routes
10.58 units through internal transfers (seven stores receive a positive
quantity), in contrast to the submitted plan. \\
\bottomrule
\end{tabularx}
\par}
\medskip
{\par\smallskip\noindent{\small\textbf{(b)} Item-\texttt{092} quantities for the eight affected stores}\par\smallskip
\scriptsize
\setlength{\tabcolsep}{3.5pt}
\renewcommand{\arraystretch}{1.05}

\begin{tabular}{@{}lrrr@{}}
\toprule
\textbf{Affected stores} & \textbf{Agent internal} & \textbf{Agent supplier} &
\textbf{Reference internal} \\
\midrule
Eight stores (aggregate) & 9.81 & 3.87 & 10.58 \\
\bottomrule
\end{tabular}
\par}
\end{table}

For example, at one affected store on 2025-03-13, the submitted plan purchased
item \texttt{ITEM\_000092} from \texttt{SUPPLIER\_DEFAULT} despite a local-store
balance of 0.0 and 327.0 units available in \texttt{WH\_LOCAL}; the same hard
gate violation occurs across the eight stores, summarized in
Table~\ref{tab:case-study-019-item}(b).

The run thus shows a rule that was asked about but not enforced.
\subsection{Audits}
\label{app:audits}

\paragraph{Audit sample and tasks.}
Two experts, Annotator A and Annotator B, independently labeled 120 clarification items (105 runs, 38 cases) and 120 extraction runs (43 cases), each sample covering all 18 configurations. Samples were drawn with fixed seeds, stratified by configuration and by coverage or recorded issue. A clarification item asks whether answering at least one recorded question would reveal the reference point (covered, uncovered, or uncertain). An extraction run compares the submitted artifact with its normalized form (\texttt{CORRECT}, \texttt{INCORRECT}, or \texttt{NOT\_ASSESSABLE}); runs without a submitted result file are \texttt{NOT\_ASSESSABLE}.

\paragraph{Post-hoc audit of simulated-customer answers.}
Annotators A and B labeled the sampled customer answers for support by the
case information and consistency with earlier turns (Table~\ref{tab:customer-answer-audit}(a)).
The audit indicates that the simulator's main weakness is cross-turn
inconsistency rather than direct conflict with the case.

\begin{table}[h]
\centering
\caption{\textbf{Audit results.}}
\label{tab:customer-answer-audit}\label{tab:human-audit}
{\par\smallskip\noindent{\small\textbf{(a)} Simulated-customer answers; counts are supported/uncertain/contradicted for support and factuality rows and not applicable/consistent/contradicted for status and consistency rows}\par\smallskip
\small
\setlength{\tabcolsep}{3.5pt}
\renewcommand{\arraystretch}{1.08}

\begin{tabularx}{\linewidth}{@{}>{\raggedright\arraybackslash}p{0.28\linewidth} r >{\arraybackslash}X >{\arraybackslash}X r r@{}}
\toprule
\rowcolor{DELIVERHeader}
\textbf{Criterion} & $n$ & \textbf{A} & \textbf{B} & \textbf{Agreement} & $\boldsymbol{\kappa}$ \\
\midrule
Answer support (all labels) & 1303 & 1250/51/2 & 1179/117/7 & 92.10\% & 0.385 \\
Answer support (non-uncertain) & 1170 & --- & --- & 99.57\% & 0.285 \\
Cross-turn status & 1303 & 968/320/15 & 965/323/15 & 99.00\% & 0.974 \\
Run factuality (all labels) & 105 & 79/24/2 & 71/28/6 & 82.86\% & 0.601 \\
Run factuality (non-uncertain) & 73 & --- & --- & 91.78\% & 0.218 \\
Run consistency & 105 & 60/36/9 & 60/35/10 & 97.14\% & 0.948 \\
\bottomrule
\end{tabularx}
\par}
\medskip
{\par\smallskip\noindent{\small\textbf{(b)} Agreement metrics between annotators and the automated five-judge panel; intervals are case-cluster bootstrap 95\% intervals for $\kappa$}\par\smallskip
\small
\setlength{\tabcolsep}{4pt}

\begin{tabularx}{\linewidth}{@{}>{\raggedright\arraybackslash}X r r r r@{}}
\toprule
\rowcolor{DELIVERHeader}
\textbf{Comparison} & $\boldsymbol{n}$ & \textbf{Agreement} &
$\boldsymbol{\kappa}$ & \textbf{95\% CI} \\
\midrule
Clarification: A--B & 114 & 96.49\% & 0.9297 & [0.8571, 0.9832] \\
Clarification: A--panel & 114 & 94.74\% & 0.8947 & [0.8054, 0.9661] \\
Clarification: B--panel & 120 & 93.33\% & 0.8667 & [0.7820, 0.9424] \\
Extraction fidelity: A--B & 120 & 91.67\% & 0.7656 & [0.5835, 0.9033] \\
\bottomrule
\end{tabularx}
\par}
\end{table}

\paragraph{Agreement and uncertainty.}
Table~\ref{tab:human-audit}(b) reports agreement and unweighted Cohen's $\kappa$ with 95\% case-cluster bootstrap intervals (5,000 resamples). Six clarification items that Annotator A marked uncertain are excluded from comparisons involving A; extraction agreement retains all three categories.

The panel does not systematically inflate Recall: its disagreements with
Annotator A are balanced between covered and uncovered labels.

\paragraph{Extraction agreement and failures.}\label{app:extraction-audit}
We audited the retained \texttt{result.json} and extractor records of all
2,646 main runs.  Numeric votes have strong majority agreement, while the
remaining non-numeric failures are rare and are broken down by configuration
in Table~\ref{tab:extraction-audit}.  Because the 120-run audit sample was
enriched for recorded extraction issues, we reweighted it to the population;
the resulting estimated extraction accuracy is 97.9\%.

\begin{table}[t]
\centering
\small
\setlength{\tabcolsep}{4pt}
\caption{\textbf{Extraction failures by configuration.}}
\label{tab:extraction-audit}
\begin{tabularx}{\linewidth}{@{}>{\raggedright\arraybackslash}X r r r@{}}
\toprule
\rowcolor{DELIVERHeader}
\textbf{Agent / model} & \textbf{Failures / slots} & \textbf{Rate} & \textbf{No artifact} \\
\midrule
\multicolumn{4}{@{}l}{\emph{\texttt{opencode}}} \\
GPT-6-Astra & 3/735 & 0.41\% & 0 \\
GLM-5.3 & 9/721 & 1.25\% & 4 \\
DeepSeek-V4-Pro & 5/731 & 0.68\% & 2 \\
Gemini-3.1-Pro & 6/733 & 0.82\% & 0 \\
Grok-4.6 & 2/733 & 0.27\% & 0 \\
Kimi-K3 & 11/731 & 1.50\% & 4 \\
Qwen3.8-Max & 10/735 & 1.36\% & 0 \\
\addlinespace
\multicolumn{4}{@{}l}{\emph{\texttt{openhands}}} \\
GPT-6-Astra & 21/729 & 2.88\% & 0 \\
GLM-5.3 & 9/711 & 1.27\% & 1 \\
DeepSeek-V4-Pro & 3/735 & 0.41\% & 0 \\
Gemini-3.1-Pro & 3/734 & 0.41\% & 1 \\
Grok-4.6 & 6/733 & 0.82\% & 0 \\
Kimi-K3 & 2/730 & 0.27\% & 3 \\
Qwen3.8-Max & 9/719 & 1.25\% & 3 \\
\midrule
\texttt{codex} + GPT-6-Astra & 1/448 & 0.22\% & 1 \\
\texttt{deepseek-harness} + DeepSeek-V4-Pro & 2/695 & 0.29\% & 0 \\
\texttt{gemini} + Gemini-3.1-Pro & 3/734 & 0.41\% & 0 \\
\texttt{kimi} + Kimi-K3 & 5/733 & 0.68\% & 2 \\
\midrule
All configurations & 110/12,820 & 0.86\% & 21 \\
\bottomrule
\end{tabularx}
\end{table}

Overall extraction failure rates are low across configurations, and majority
voting reduces the effect of individual extractor disagreements on final scores.

\FloatBarrier
\subsection{Prompt templates and interaction interfaces}
\label{app:prompts}

This section reproduces the benchmark prompts for the solving agent,
simulated customer, extractor, and judge, and the question--answer
interfaces.  They form the benchmark layer added to each scaffold; the
scaffolds' own prompts are unchanged.  Double-braced fields are filled at
runtime.  The listings below show the exact shared and role-specific
templates used in the benchmark, including the information boundaries and
the no-repair extraction contract.  Scaffold-specific wrappers and exact
field substitutions are released with the artifact in
\path{analysis/prompt_templates/}.

\begingroup
\lstdefinestyle{deliverprompt}{
  language={},basicstyle=\ttfamily\fontsize{8.5}{10.5}\selectfont,
  columns=fullflexible,keepspaces=true,breaklines=true,breakatwhitespace=true,
  showstringspaces=false,frame=none,numbers=none,
  aboveskip=0pt,belowskip=0pt}
\lstset{style=deliverprompt}
\ifdefined\paperstyle
\tcbset{deliverprompt/.style={
  enhanced jigsaw,breakable,
  colback=DELIVERPromptBack,colframe=DELIVERPromptFrame,
  colbacktitle=DELIVERPromptTitle,coltitle=black!85,
  fonttitle=\rmfamily\bfseries\small,
  boxrule=0.4pt,arc=1pt,
  left=8pt,right=8pt,top=7pt,bottom=7pt,
  toptitle=4pt,bottomtitle=4pt,
  before skip=10pt,after skip=12pt,
  pad at break*=3pt,lines before break=4}}
\newenvironment{deliverpromptbox}[1]{%
  \begin{tcolorbox}[deliverprompt,title={#1},
    title after break={#1\hfill\normalfont\rmfamily\footnotesize(continued)}]}{%
  \end{tcolorbox}}
\else
\newenvironment{deliverpromptbox}[1]{%
  \par\noindent\arrayrulecolor{DELIVERPromptFrame}%
  \begin{tabular}{|p{1.3\linewidth}|}\hline
  \rowcolor{DELIVERPromptTitle}\textbf{#1}\\\hline
  \rowcolor{DELIVERPromptBack}}{%
  \\\hline\end{tabular}\par}
\fi

\subsubsection{Solving-agent instructions}
\label{app:agent-prompts}
All conditions use the same base template.  The request block is the heading
\texttt{\#\# The client's request} followed by the complete
\texttt{instruction.md}; the data list contains the sorted filenames
directly under \texttt{data/}.  The workspace is the run's solving directory.
Only the condition block changes.

\begin{deliverpromptbox}{Shared agent template}
\begin{lstlisting}[style=deliverprompt]
You are an algorithm delivery engineer. A client has sent you a business request and some data files.

Your job: understand what they actually need, then build a working solution and deliver it.

## Working directory

{{workspace}}

Data files provided by the client:
{{data_files}}

Write all outputs to the working directory.

{{request_block}}

{{condition_block}}

## Delivering

Do the work -- do not merely describe how you would do it. The task is complete only
when the deliverable files exist in the working directory.

Produce whatever artifact the request calls for. The client did not specify a file
format -- infer what is appropriate from what they asked for.

Also write a short `NOTES.md` recording your key modelling decisions: what you
assumed, what you inferred from the data, and anything you were unsure about.
\end{lstlisting}
\end{deliverpromptbox}
\begin{deliverpromptbox}{Hidden condition block}
\begin{lstlisting}[style=deliverprompt]
## Missing information

The request is written in the client's own words. It will not spell out every
detail you need -- data conventions, units, edge-case handling and similar points
are often left implicit because the client considers them obvious.

There is no one available to answer questions. Where something cannot be
determined from the request or the data, make a reasonable assumption, proceed,
and record the assumption in `NOTES.md`.
\end{lstlisting}
\end{deliverpromptbox}
\begin{deliverpromptbox}{Interact condition block}
\begin{lstlisting}[style=deliverprompt]
## Missing information

The request is written in the client's own words. It will not spell out every
detail you need -- data conventions, units, edge-case handling and similar points
are often left implicit because the client considers them obvious.

The client is reachable and expects to be consulted on requirements. If something
cannot be determined from the request or the data, ask them. You may ask more
than once, and at any point in the work.

When you ask, put the questions at the end of your reply under a heading
`## Questions for the client`, one question per line, and stop there -- the client
will answer in the next message. (Asking in plain prose also works; the heading
just makes the exchange cleaner.)

For anything you choose not to ask about, make a reasonable assumption, proceed,
and record the assumption in `NOTES.md`.
\end{lstlisting}
\end{deliverpromptbox}
\begin{deliverpromptbox}{Interact-Req condition block}
\begin{lstlisting}[style=deliverprompt]
## Missing information

The request is written in the client's own words. It will not spell out every
detail you need -- data conventions, units, edge-case handling and similar points
are often left implicit because the client considers them obvious.

The client is reachable. Before writing any code, ask the client about anything
you need settled, and keep asking until you have what you need. Do not start
the work until then.

Ask about anything the request and the data do not settle. You may ask across
several rounds.

When you ask, put the questions at the end of your reply under a heading
`## Questions for the client`, one question per line, and stop there -- the client
will answer in the next message. (Asking in plain prose also works; the heading
just makes the exchange cleaner.)

Proceeding on an unconfirmed reading counts as a failure even if the guess
turns out right. Record in `NOTES.md` any assumption the client did not settle.
\end{lstlisting}
\end{deliverpromptbox}

\begin{deliverpromptbox}{Full condition block}
\begin{lstlisting}[style=deliverprompt]
## Additional context

The client has already supplied the following details, covering the data
conventions, units and edge-case handling that the request itself leaves implicit:

{{information}}

There is no one available to answer further questions. Where something is still
undetermined, make a reasonable assumption, proceed, and record it in `NOTES.md`.
\end{lstlisting}
\end{deliverpromptbox}
\begin{deliverpromptbox}{Oracle answer-coverage condition block}
\begin{lstlisting}[style=deliverprompt]
## Missing information

The request is written in the client's own words. It will not spell out every
detail you need -- data conventions, units, edge-case handling and similar points
are often left implicit because the client considers them obvious.

The client has already settled the following points for you:

{{answers}}

There is no one available to answer further questions. Where something still
cannot be determined from the request, the data, or the points above, make a
reasonable assumption, proceed, and record the assumption in `NOTES.md`.
\end{lstlisting}
\end{deliverpromptbox}

For \texttt{Full}, \texttt{information} is the complete business-information
file.  For the oracle experiment, \texttt{answers} is the selected answer
text, rendered as bullets in registry order, or \texttt{(none)} for zero
coverage.  The nested subset selection is specified in
Appendix~\ref{app:information-ablation}; requirement identifiers and type
labels are not inserted into the oracle prompt.

\subsubsection{Simulated-customer messages}
\label{app:customer-prompt}
The customer uses GLM-5.3 and the default answering style below.  Each call
has this system message, the accumulated user-question/assistant-answer
history, and the current question as a new user message.  The returned
answer is appended to that history.  The two knowledge fields are the
initial request and complete business information; clarification annotations
and evaluator code are not included.  The retained implementation requests
at most 10,000 output tokens per call and leaves temperature and seed unset.

\begin{deliverpromptbox}{Customer system prompt (English translation)}
\begin{lstlisting}[style=deliverprompt]
You are the person responsible for this business scenario, answering the technical team's clarification questions about the requirements.
The following is your business knowledge. Before answering, read each block in full and cross-check them. In particular, <information> often contains specific values, field names, coordinates, thresholds, and constraint details. When asked about a schema, parameter, threshold, coordinate, formula, or similar detail, first look for the existing answer in <information>; do not say "I will provide it" or decide it on the spot:

<instruction>
{{instruction}}
</instruction>

<information>
{{information}}
</information>

Note: Any of the above may contain internal technical details such as evaluation formulas, scoring weights, or penalty coefficients. Do not volunteer these to the technical team. If asked, you may provide a qualitative business-level explanation, but do not give specific mathematical formulas or numerical weights.

Mandatory answering procedure (follow these three steps for every question):

Step 1: Check information first (internally; do not include this step in the reply).
Before selecting or writing an answer, find the corresponding text in <information> and answer on that basis.
You are playing the business owner: this is business knowledge in your head, not a document you can consult.
Do not say "checking information", "according to the document/materials/instructions", "original excerpt", or "not found", or imply that you are consulting materials. Answer directly in your own voice.

Step 2: Compare the options with the facts in information.
- If an option exactly matches the facts in information, select it.
- If none of the options exactly matches, select "Other (please describe)" and copy the ranges, values, or field names from information.
- Only if information does not address the question may you decide based on instruction.

Step 3: Do not replace specific values with vague qualifications.
Do not answer "unrestricted / not counted / unnecessary / flexible / for the model to decide" unless information explicitly says so. Copy specific values, ranges, and field names from information.

If selecting an option that says "please describe", provide the specific content after the option name. If there is nothing new to add, select a negative option rather than leaving the description empty.

Answer requirements:
- Speak as the business owner. Give only the answer itself, without explaining the reasoning or repeating the question.
- If there are multiple questions, answer each directly on a separate line.
- Do not volunteer implementation details such as evaluation methods, scoring formulas, or exact objective weights. If asked, you may give a qualitative business-level explanation, but not specific mathematical formulas or numerical weights.
- If none of the information sources covers the question, provide a reasonable business default.
\end{lstlisting}
\end{deliverpromptbox}

This prompt allows business defaults for unsupported questions, so answers
are not guaranteed to follow from the case information
(Appendix~\ref{app:audits}).

\subsubsection{Question and answer interfaces}
\label{app:question-interface}
The harness supports native question requests, a scaffold callback, and
text exchanges.  These routes share the customer above but preserve each
scaffold's interaction mechanism.  The following examples illustrate the
message shapes; angle-bracket values are placeholders, not recorded dialogue.

\paragraph{OpenCode native question tool.}
The harness receives a request containing a \texttt{questions} array and
replies through \texttt{POST /question/<request-id>/reply} (or the
corresponding session-scoped route in the older API).  Each answer is a
one-element string array, preserving free-text answers rather than forcing
them into an agent-proposed option.  The customer receives each
\texttt{question} string, falling back to \texttt{header} if it is empty;
option metadata is recorded but is not separately appended to that string.

\begin{deliverpromptbox}{Native question request and reply}
\begin{lstlisting}
Request:
{"id": "<request-id>", "questions": [
  {"question": "<question text>", "header": "<short header>",
   "options": [{"label": "<option>", "description": "<meaning>"}]}
]}

Reply:
{"answers": [["<customer answer>"]]}
\end{lstlisting}
\end{deliverpromptbox}

\paragraph{Text exchange and OpenHands callback.}
The CLI configurations use text exchanges.  OpenHands obtains a reply via
its user-response callback, connected to the customer by the harness.
Questions under the suggested heading are parsed directly when possible;
otherwise a detector determines whether the message requests information.
Each detected question is answered separately, and the answers return as
one user message in the form below.  Recorded question units used for
scoring may be split further as described in Appendix~\ref{app:process-metrics}.

\begin{deliverpromptbox}{Text-channel request and response}
\begin{lstlisting}
Agent message:
## Questions for the client
<question 1>
<question 2>

Customer reply delivered as a user message:
- <question 1>
  <answer 1>
- <question 2>
  <answer 2>
\end{lstlisting}
\end{deliverpromptbox}

The detector sees only agent text, not hidden requirements, and is skipped
when the question heading is present.  Replies without a question mark are
treated as non-questions; otherwise five models vote by strict majority of
returned ballots on whether the agent is asking.  Overlong replies keep the
first 2,000 and last 8,000 characters.

\begin{deliverpromptbox}{Question detector system message}
\begin{lstlisting}[style=deliverprompt]
You judge whether a coding agent's reply is ASKING the client for information, or REPORTING/DELIVERING work.

Return JSON only:
{"asking": true|false, "questions": ["...", "..."]}

Rules:
- "asking" is true only if the agent is waiting for the client to supply
  information it cannot determine itself.
- Rhetorical questions inside an explanation or report are NOT asking.
  ("How do we balance time against fairness? We weight them 1:90." -> false)
- Offers to continue are NOT asking. ("Shall I also plot the routes?" -> false)
- Split compound questions into atomic ones, one fact per entry.
  ("What speed, and is loading time included?" -> two entries)
- Keep each question in its original language, close to the agent's wording.
- If "asking" is false, "questions" must be [].
\end{lstlisting}
\end{deliverpromptbox}
\begin{deliverpromptbox}{Question detector user message}
\begin{lstlisting}[style=deliverprompt]
Agent reply:

{{clipped_agent_reply}}{{hint}}
\end{lstlisting}
\end{deliverpromptbox}

The detector requests at most 2,000 output tokens.  On malformed JSON it
retries once, prepending the following text to the original user message:
\begin{deliverpromptbox}{Question detector JSON-repair retry prefix}
\begin{lstlisting}
Your previous response violated the JSON protocol (ValueError). Return exactly one JSON object with keys asking and questions, and no other text.
\end{lstlisting}
\end{deliverpromptbox}

The retained text loop sends the following continuation when an agent has
stopped without a deliverable, subject to its run budget, and allows up to
three such pushes in that branch.  Repeated identical question batches or
exhausted clarification limits also trigger a transition to solving.

\begin{deliverpromptbox}{Continuation and budget messages}
\begin{lstlisting}
Continuation:
Understood. Please now carry out the work and write the deliverable files into the working directory.

Text-channel clarification budget exhausted:
(No further questions can be answered. Proceed with your best judgement.)

Native-tool / callback clarification budget exhausted:
No further information is available. Proceed with your best judgement.
\end{lstlisting}
\end{deliverpromptbox}

If a customer call fails, the OpenCode handler returns a placeholder telling
the agent to proceed on its own judgment.

\subsubsection{Artifact-extraction instructions}
\label{app:extractor-prompt}
The extractor runs after the solving agent has finished.  The shared prompt
below inserts the case's complete \texttt{submission\_schema.json} under
\texttt{Target schema (authoritative)}.  If that file is absent, it uses the
evaluator's \texttt{EXTRACTOR\_SPEC} documentation, or directs the extractor
to infer the schema from evaluator source.  Paths identify the retained
agent workspace, case data, evaluator, and a separate normalized-output
directory.  
\begin{deliverpromptbox}{Shared extractor template}
\begin{lstlisting}[style=deliverprompt]
You are a data extractor for a benchmark evaluation pipeline.

## Your task

1. **Read `_agent_summary.md` first** (if present in the workspace). It tells you which file
   the agent considers its final answer. The agent's own designation decides which artifact
   gets graded -- you do not get to pick a different (better-scoring) one.
2. Read the target schema below to learn what files and fields the evaluator accepts.
3. Find the agent's output in the workspace. Filenames and column names will differ.
4. Normalize it to the schema and write the result to the OUTPUT directory.
5. **If the deliverable is executable code** (the schema asks for a `.py` defining a specific
   function or class): the evaluator imports it from the OUTPUT directory, so copy the entry
   file under the required name AND every workspace-local module it imports, directly or
   transitively. Copying only the entry file leaves it orphaned -- ModuleNotFoundError, and a
   working solution scores 0.
6. Write `_result.json` to the OUTPUT directory as your final action.

{{schema_block}}

{{verification_block}}

**Stop as soon as the evaluator returns a score.** A 0 is a legitimate final answer: the
agent's solution ran and did not meet the constraints. Never edit values, add rows, drop
rows, or swap in a different solution file to raise the number. You normalize
*representation* -- filenames, column names, layout, encoding. The decisions are the agent's
and must survive unchanged. If no score can be obtained, the artifact is wrong; say so and
stop rather than making something up.

**Forbidden transformations.** The following transformations would change the
agent's substantive answer and are therefore forbidden.

1. **Nudging values inside a stated tolerance.** Do not adjust a number within an evaluator
   tolerance (or any other slack) so that a failing constraint passes. Tolerance exists to
   absorb *the agent's* rounding, not to give the extractor a search space. If the agent's
   numbers land outside a limit, that is the answer: report the 0.
2. **Writing an adapter when the agent's interface does not match the evaluator's.** When a
   deliverable is code and its signature or data model disagrees with what the evaluator
   passes in (e.g. the policy expects a custom object, the evaluator hands it a dict), do
   **not** author a shim, a wrapper class, or a re-implementation. Copy the agent's file
   as-is, let it fail, and record the mismatch in `notes`.
3. **Dropping rows that "look wrong".** Removing rows the agent emitted -- even ones that
   seem like annotations, placeholders, or infeasible leftovers -- changes the answer. If you
   drop or add **any** row, state the before/after counts and the reason in `notes`.

4. **"Fixing" an identifier the agent spelled wrong.** Whether you may repair a bad name
   depends on one question: **does the task specify the naming?**
   - **Task is silent on naming** (e.g. the agent labels a plate `plate1` where the evaluator
     keys on `P001`): map it, provided the target is **unique** in the reference data. This
     is ordinary identifier normalization. If the mapping is ambiguous or has no match,
     do not guess -- leave it and say so.
   - **Task states the naming rule** (e.g. `information.md` says feature names must be
     real columns of the training data, in `{trait}_D{day}` form): a name that breaks the
     rule is a **substantive error by the agent**, not a formatting slip. Copy it as-is and
   let it score 0.
   Check `information.md` / `instruction.md` for a naming rule **before** you repair any
   name. Also check the schema's `role`: repairing is only ever in scope for `identifier`
   fields, never for `decision` fields -- those are the answer itself.

5. **Using the evaluator's own formula to generate a value the evaluator checks.** When the
   schema documents how a field relates to others (`sub_i_long must equal
   mother_cut_weight / (mother_width x 7.65) x 10000`), that formula is there so you can
   understand the field -- **not so you can compute it**. Filling the field from the formula
   makes the consistency check pass by construction: the evaluator ends up verifying its own
   arithmetic instead of the agent's. A derived value that the evaluator validates is
   load-bearing even when it is not itself a decision variable.
   The test is not "is this field a decision?" -- it is **"does the evaluator look at it?"**
   If yes, the value must come from the agent, even when you can compute a better one, and
   even when the agent's is provably inconsistent. An inconsistency you found is a finding to
   report in `notes`, not a defect to repair.

## When the agent's designated file is incomplete

Rule 1 says the agent's own designation decides what gets graded. That assumes the
designated artifact exists and is coherent. When it is **not**:

- **The designation names a state the agent never finished.** If the designated artifact is
  incomplete or internally contradictory, fall back to the most recent complete state that
  passes the agent's own validation, set `status` to `partial`, and say in `notes` which state
  you used and why the designated one was left out. This is not "picking the better-scoring
  file" -- it is selecting the only file that is a coherent answer.
- **Extra rows beyond the graded set.** Rows the agent produced for keys that are not in the
  evaluation set should be dropped, with the before/after counts disclosed. That is different
  from dropping rows the agent *did* produce for graded keys, which is forbidden.

## Verify order-bearing fields before you write

When a field encodes a sequence (a visit order, a permutation, or a concatenated route), its
order **is** the decision. Reconstructing it from a grouped source can silently scramble the
group order even when all members and counts match.

Before writing an order-bearing field, re-derive it from the source a second way and compare
element by element. If the schema gives you a companion structure (a `trips` array with per-
trip stop counts, say), slice your sequence by those counts and check each slice against the
source group it should match.

Rule of thumb: if an edit would change the score, it is not normalization. Reformatting,
renaming, reshaping, re-encoding never change a score -- those are yours to make freely.

## Normalization rules

- **Semantic-first mapping**: the same column name does not guarantee the same meaning.
  Check the schema's field definition (units, scope, total vs. net) before renaming.
- **Column renaming / format conversion**: fuzzy-match agent columns to schema fields;
  convert wide/long layouts as needed.
- **Missing rows**: only fill a missing combination with zero when the schema defines that
  dimension's absence AS zero. Do NOT invent rows for tasks the agent did not produce --
  leave them out and set status to "partial".
- **Never** re-run the agent's solver, re-optimize, or re-compute its decisions.
- **Prefer the highest-precision artifact.** When the agent expresses the same decisions in
  several files -- say a minute-truncated CSV and a full-precision JSON -- take the lossless
  one, even if `_agent_summary.md` names the other. Precision loss is not a formatting
  choice; it silently changes the answer being graded.
- **Never take anything from the Data directory as the agent's output.** `data/` holds the
  problem input, including any baseline or seed policy shipped with the task. Submitting one
  of those grades the benchmark's own reference instead of the agent -- if the workspace has
  no real deliverable, say so and set status to "failed".

## Paths & permissions

| Directory | Path | Read | Write |
|-----------|------|------|-------|
| Workspace | `{{workspace}}` | YES | **NO** |
| Data      | `{{data_dir}}` | YES | **NO** |
| Evaluator | `{{evaluator}}` | YES | **NO** |
| Output    | `{{output_dir}}` | YES | **YES** |

All writes go to the Output directory only.

## _result.json format

{
  "status": "success",
  "notes": "what was found and what transformations were applied",
  "files_found": ["relative paths of agent output files discovered"],
  "files_normalized": ["filenames written to the output directory (excluding _result.json)"]
}

Status: "success" = all required files written; "partial" = found output but it is
incomplete or self-contradictory (describe what is missing); "failed" = no relevant agent
output found. When the agent's output is genuinely incomplete, "partial" is the correct
outcome -- do NOT reconstruct or re-solve to reach "success".
\end{lstlisting}
\end{deliverpromptbox}

The \texttt{verification\_block} uses the first block below for ordinary
cases.  Case 13, whose evaluator launches expensive solver subprocesses,
uses the second block and is scored after the extractor exits.

\begin{deliverpromptbox}{Standard extractor verification block}
\begin{lstlisting}[style=deliverprompt]
## Verify by running the evaluator (required)

A schema cannot enumerate every shape an agent might produce, so do not guess whether your
normalization is right -- check it:

```bash
python3 {{evaluator}} --data-dir {{data_dir}} --submission-dir {{output_dir}}
```

React to *structural* failures only:

| What you see | What it means | What to do |
|---|---|---|
| `ModuleNotFoundError` | a module the entry file imports is missing | copy that module in too |
| `AttributeError: must define <fn>` | wrong entry file | try the next candidate |
| `missing <file>` / `KeyError` / missing column | filename or column mapping is off | fix the mapping |
| valid JSON with any score, **including 0** | the evaluator ran to completion | **stop -- you are done** |
\end{lstlisting}
\end{deliverpromptbox}
\begin{deliverpromptbox}{Executable-case verification exception (case 13)}
\begin{lstlisting}[style=deliverprompt]
## Heavyweight executable case

Do not run `{{evaluator}}` or the Agent solver. Copy the selected Agent Python
source byte-for-byte to the required output filename. Do not rewrite its CLI,
output shape, algorithm, or decisions: incompatibility is an Agent result.
The scoring stage evaluates it after you exit.
\end{lstlisting}
\end{deliverpromptbox}

The extractor's \texttt{status} describes the extraction only; the
evaluator alone scores the artifact (Appendix~\ref{app:delivery-metrics}).

\subsubsection{Clarification-coverage judge messages}
\label{app:judge-prompt}
Each judge receives the same system message and, for each registered
requirement, the user template below.  The numbered question list is the
complete run-level list after splitting.  Calls use temperature zero.

\begin{deliverpromptbox}{Coverage judge system message}
\begin{lstlisting}[style=deliverprompt]
You grade whether an agent's clarification questions uncovered a specific hidden requirement. Judge strictly and reply with JSON only.
\end{lstlisting}
\end{deliverpromptbox}
\begin{deliverpromptbox}{Coverage judge user message}
\begin{lstlisting}[style=deliverprompt]
A client holds the hidden requirement below. An agent asked the questions
below to clarify the task. Decide what the client would reveal in reply.

HIDDEN REQUIREMENT (id: {{requirement_id}})
{{requirement_description}}
{{optional_trigger_block}}

AGENT'S QUESTIONS (each on its own line, numbered)
{{numbered_questions}}

The client answers cooperatively and concretely: when a question points at a
topic they hold information about, they state the actual values or rules rather
than a generic description. They do NOT volunteer information the agent never
asked about.

Judge two things:

1. ATTEMPTED -- does any question point at this requirement's topic, however
   loosely phrased?
2. COVERED -- answering that question, would the client state the specific
   requirement above?
   - Yes if the question names the specific quantity, rule, field or entity this
     requirement is about, so the client knows exactly which convention to state.
     The agent need not already know the value, nor use the same wording; asking
     what the convention is counts.
   - No if the question merely names a topic, module or data source and asks what
     to "watch out for", what the "rules" are, or what is "special" -- even when
     that topic is the right one. Such a question does not tell the client which
     of the many conventions in that area is wanted, so a cooperative client
     would ask back rather than recite this requirement.
     Examples that must be judged No: "anything special about the data?",
     "any rules I should know about billing?", "any pitfalls when reading these
     sheets?". Naming the right domain does not make a question specific.

Reply with JSON only:
{"attempted": true|false, "covered": true|false,
  "question_indices": [<1-based indices that cover it>]}
\end{lstlisting}
\end{deliverpromptbox}
\begin{deliverpromptbox}{Optional trigger-example block}
\begin{lstlisting}[style=deliverprompt]

QUESTIONS THAT WOULD SURFACE IT (reference examples, not exhaustive --
a differently-worded question that targets the same fact also counts)
{{indented_trigger_bullets}}
\end{lstlisting}
\end{deliverpromptbox}

Only \texttt{covered} enters Recall, and the returned indices determine
Precision.  A malformed reply is retried once with the prefix below.

\begin{deliverpromptbox}{Coverage judge JSON-repair retry}
\begin{lstlisting}[style=deliverprompt]
Your previous response violated the JSON protocol (ValueError). Return exactly one JSON object matching the required schema. Do not include analysis, Markdown, or any text before or after it.

{{original_judge_user_message}}
\end{lstlisting}
\end{deliverpromptbox}
\endgroup

\end{document}